\documentclass{tADR2e}

\usepackage{url}
\usepackage{hyperref}
\usepackage{subcaption}
\renewcommand{\thesubfigure}{(\alph{subfigure})}
\usepackage{multirow}
\usepackage{booktabs}
\usepackage{bbm}
\usepackage[anythingbreaks]{breakurl}
\usepackage{amssymb}
\usepackage{algorithmic}
\usepackage{algorithm}

\newcommand{\myrot}[1]{\rotatebox{90}{#1}}
\usepackage[nobottomtitles]{titlesec}

\DeclareMathOperator*{\argmax}{arg\,max}
\DeclareMathOperator*{\argmin}{arg\,min}

\begin{document}

\jvol{00} \jnum{00} \jyear{2013} \jmonth{January}

\articletype{REGULAR}

\title{Multi-source Pseudo-label Learning of Semantic Segmentation \\
	for the Scene Recognition of Agricultural Mobile Robots}

\author{S. Matsuzaki$^{a}$$^{\ast}$\thanks{$^\ast$Corresponding author. Email: matsuzaki@aisl.cs.tut.ac.jp \vspace{6pt}},
		J. Miura$^{a}$,
		and H. Masuzawa$^{a}$\\\vspace{6pt}  $^{a}${\em{Department of Computer Science and Engineering,
				Toyohashi University of Technology, Toyohashi, Japan }}
}

\maketitle

\begin{abstract}
	This paper describes a novel method of training a semantic segmentation model for scene recognition of agricultural mobile robots exploiting multiple publicly available datasets that are different from the target greenhouse environments.  Semantic segmentation models require abundant labels given by tedious manual annotation for training.  Although unsupervised domain adaptation (UDA) is studied as a workaround for such a problem, existing UDA methods assume a source dataset similar to the target dataset, which is not available for greenhouse scenes.  In this paper, we propose a method to train a semantic segmentation model for greenhouse images leveraging multiple publicly available  datasets not dedicated to greenhouses.  We exploit segmentation models pre-trained on each source dataset to generate pseudo-labels for the target images based on agreement of all the pre-trained models on each pixel.  The proposed method allows for effectively transferring the knowledge from multiple sources rather than relying on a single dataset and realizes precise training of semantic segmentation model.  We also introduce existing state-of-the-art methods to suppress the effect of noise in the pseudo-labels to further improve the performance.  We demonstrate that our proposed method outperforms existing UDA methods and a supervised SVM-based method.

	\medskip
\end{abstract}

\section{Introduction}

Semantic segmentation based on Deep Neural Network (DNN) is
widely used in the scene recognition system
of self-driving vehicles and autonomous mobile robots.
In our research project, we are aiming at developing an agricultural robot
that can recognize regions covered by traversable plants and go through them
\cite{Matsuzaki2018,Matsuzaki2022}.
For such an application, semantic segmentation is suitable
since it provides pixel-wise semantic information.
Usually, DNNs are trained using a large amount of
data with hand-annotated labels to ensure high accuracy and generalization.
However, 
hand-annotation is time consuming and physically demanding.
In the case of introducing a mobile robot in new environments,
it is not realistic to prepare such a fully labeled dataset
for every environment even though 
a model should be trained specifically for the environment.

To address this problem, one might consider applying a model trained with
existing publicly available image datasets with ground truth labels.
The performance of pre-trained models, however, often deteriorates
on different data domains
due to the difference in data distributions
between the training dataset and the target scenes,
known as ``domain shifts'' \cite{Csurka2017}. 
Domain adaptation (DA) is a task to adapt a model trained on
one data domain 
to another domain with minimal degradation of performance.
Unsupervised domain adaptation (UDA) is a type of DA
where labeled source datasets and unlabeled target datasets are available.
UDA can mitigate the burden of hand-annotation
by leveraging information in the source domain and
transferring it to the target.

A major task of UDA is adaptation from labeled synthetic images
automatically acquired in a simulator to unlabeled real images.
The adaptation task from GTA 5 \cite{Richter2016} or SYNTHIA \cite{Ros2016}
to Cityscapes \cite{Cordts2016} is one of the most widely used benchmarks and
many studies have verified their effectiveness on these tasks.
However, it is often difficult to prepare appropriate source datasets with
full segmentation labels for target environments in the wild.
In the case of greenhouses, there is no real image dataset of greenhouses
with segmentation labels, and rich simulation environments
applicable to our task.

Instead, we consider utilizing publicly available
rich image datasets of the real world not specifically designed for greenhouses.
\begin{figure}[tb]
	\begin{minipage}[t]{0.33\linewidth}
		\centering
		\includegraphics[width=\hsize]{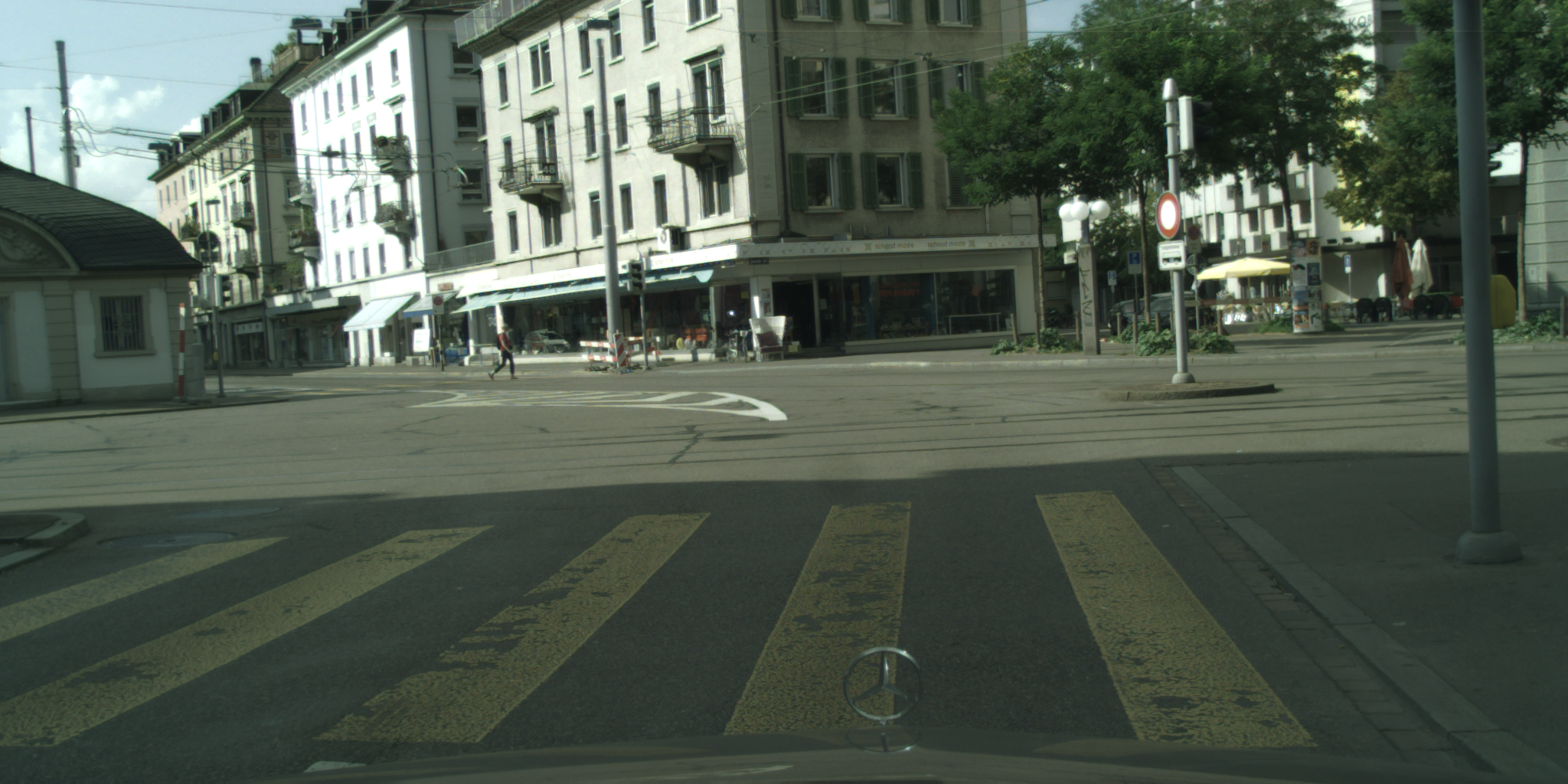}
		\subcaption{Cityscapes \cite{Cordts2016}}
		\label{fig:difference_of_urban_and_greenhouse_urban}
	\end{minipage}
	\begin{minipage}[t]{0.3\linewidth}
		\centering
		\includegraphics[width=\hsize]{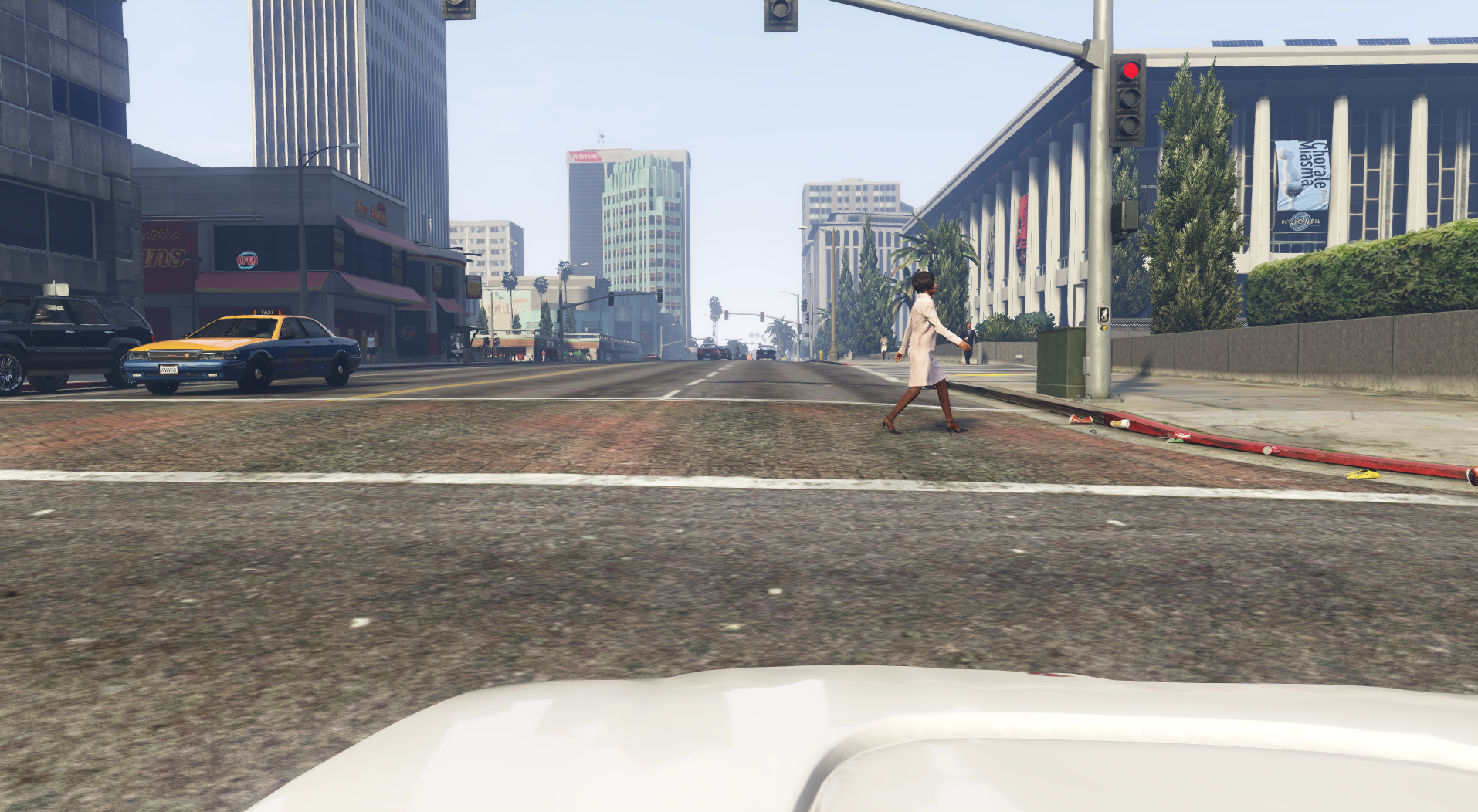}
		\subcaption{GTA5 \cite{Richter2016}}
		\label{fig:difference_of_urban_and_greenhouse_gta5}
	\end{minipage}
	\begin{minipage}[t]{0.3\linewidth}
		\centering
		\includegraphics[width=\hsize]{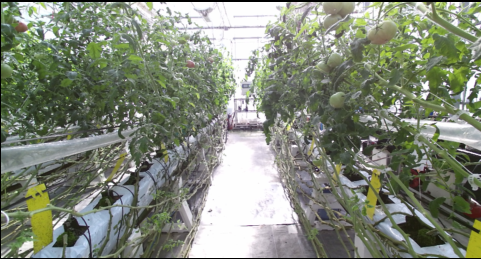}
		\subcaption{Greenhouse}
		\label{fig:difference_of_urban_and_greenhouse_greenhouse}
	\end{minipage}
	\caption{Difference of appearance between an urban scene and a greenhouse}
	\label{fig:difference_of_urban_and_greenhouse}
\end{figure}
Fig. \ref{fig:difference_of_urban_and_greenhouse} shows
real and synthetic urban scenes as well as a greenhouse scene
for comparing the difference of appearance.
The difference between real and synthetic urban scenes
(Figs. \ref{fig:difference_of_urban_and_greenhouse_urban} and
\ref{fig:difference_of_urban_and_greenhouse_gta5})
mainly stems from the discrepancy of their characteristics of appearance
such as textures.
However, they share a similar distribution of the objects in the images,
e.g., the majority of areas are occupied by buildings and roads.
On the other hand, the images of real urban scenes and the greenhouses
(Figs. \ref{fig:difference_of_urban_and_greenhouse_urban} and
\ref{fig:difference_of_urban_and_greenhouse_greenhouse})
are both real, but there are instead
structural differences such as the distribution and
scales of the objects in the images, etc.
These structural differences of the datasets
makes the existing UDA methods less effective.
In fact, as later shown in Sec.
\ref{sec:result-result-comparison},
state-of-the-art UDA methods 
result in poor performance.

While the outputs of the source models contain a lot of noise,
we also observed that all the models provide accurate classification on large areas
of the images (more details are in Sec. \ref{sec:pseudo-label_generation}).
Based on the observation, we propose to exploit multiple datasets
as source datasets for pseudo-label generation,
rather than relying on a single dataset dissimilar to the target dataset.
Our key contribution is the method to generate pseudo-labels
using models trained with multiple publicly available datasets
to effectively transfer knowledge to structurally different target images 
without ground truth labels.
Pseudo-labels for the target images are generated
by merging outputs from multiple models,
each of which is pre-trained with a labeled source dataset.
In particular, a pseudo-label is assigned on a pixel only when
all the source models predict the pixel as the same object class.
This method effectively filters out misclassified labels
and leads to a better performance of the trained models.
To further suppress the effect of noise in the pseudo-labels,
we also introduce a loss weighting based on
the uncertainty of estimation and
pseudo-label denoising inspired by Zheng et al. \cite{Zheng2021}
and Zhang et al. \cite{Zhang2021}, respectively.

We demonstrate the effectiveness of the method in the experiments
of adaptation from three source datasets of urban or outdoor scenes,
to multiple greenhouse datasets as the target.
%


\section{Related work}\label{related-work}
\hypertarget{semantic-segmentation}{%
\subsection{Semantic segmentation}\label{semantic-segmentation}}
\subsubsection{DNN architecture}
Semantic segmentation is a task to predict an object class on each pixel
of an image. It is one of the most important tasks in the field of
computer vision and its application includes scene recognition for
the navigation of self-driving cars and autonomous mobile robots. 
After Long et al. \cite{Long2015} proposed the idea of converting classification networks
into a fully convolutional network to produce a probability map for the input of
arbitrary size, most of the semantic segmentation models have been
developed based on it \cite{Badrinarayanan2017,Ronneberger2015}. 
Many recent high accuracy models use ResNet proposed by He et al. \cite{He2016} 
as a backbone network to build very deep networks \cite{Zhao2017a,Chen2018}.
While it is proven to be effective to stack many layers to get a high accuracy,
the performance comes at the high computational cost.
To tackle this problem, computationally efficient models have also been studied,
targeting the applications such as robotics and autonomous driving cars where
real time recognition is required for the operation
\cite{Mehta2018,Mehta2019,Paszke2016,Romera2018}.
\subsubsection{Semantic segmentation for navigation in agriculture}
Image segmentation techniques have been studied for navigation tasks in agricultural fields.
Lulio et al. \cite{Lulio2009} proposed an image segmentation method 
for robot navigation in orchards with a combination of color-based region segmentation
and an artificial neural network.
Sharifi et al. \cite{Sharifi2015},
Aghi et al. \cite{Aghi2021}, and Lin et al. \cite{Lin2019} proposed 
DNN-based segmentation methods for the purpose of robot navigation.

In contrast to ``shallow'' methods, the DNN-based methods have several advantages.
First, the generalization performance of DNNs is better than that of
the image processing methods and machine learning methods with hand-crafted features.
Second, the trained DNNs provide discriminative ``deep'' features
that can be used in other tasks \cite{Yang2015a}.
In fact, we used the features learned in the proposed method 
to estimate the \textit{traversability} of the objects 
and showed that the features are capable of discriminating between 
traversable and non-traversable plants \cite{Matsuzaki2022}.

Moreover, unlike other DNN-based methods, 
our proposed methods do not require any image datasets with manually provided labels 
dedicated for the target environment, i.e., greenhouses.

%
%

\subsection{Domain adaptation for semantic segmentation}
\label{Domain-adaptation}

\subsubsection{Unsupervised domain adaptation}
Semantic segmentation based on deep learning usually requires a large
amount of training data labeled on each pixel and manual labeling is especially
time-consuming. 
One promising approach is
to use a large amount of image-label pairs automatically synthesized by computer graphics. 
Since the appearance of the synthetic images is different from real-world images,
the model trained on the synthetic data does not perform well in
the real world. Therefore, we need UDA to re-train the
model in a different domain i.e., the real environment. 
There are two major approaches to UDA methods \cite{Zhao2020,Wilson2020}.
One is called ``domain alignment'' where a model is trained
so that the distribution discrepancy of domains is minimized.
The alignment is done in different levels such as 
the input space \cite{Hoffman2018,Vu2019}, feature space \cite{Saito2017,Ganin2017a}, 
and output label space \cite{Tsai2018,Biasetton2019}.
Those methods aim to learn domain-invariant features of both domains.
Recently, such training is often realized by adversarial training \cite{Goodfellow2014,Ganin2017a}.

The other approach of UDA is pseudo-label-based methods.
This type of method uses outputs from a model pre-trained
with a source dataset as true labels.
By directly training the model on the target images 
with the pseudo-labels, it can fully learn
domain-specific information.
However, the pseudo-labels for the target images inevitably
contain misclassification and it may affect the training as noise.
To deal with the noise, different approaches have been 
proposed, such as thresholding based on the confidence of the prediction \cite{Zou2018,Zou2019},
the adaptive weighting of loss values based on the uncertainty of the prediction \cite{Zheng2021},
and prototype-based pseudo-label denoising \cite{Zhang2021}.

\subsubsection{Multi-source domain adaptation}
Apart from the aforementioned single-source adaptation methods,
there are several studies of multi-source domain adaptation (MDA).
Mansour et al. stated that the target distribution can be represented as
a weighted mixture of multiple source distributions.
Following this analysis, Xu et al. \cite{Xu2018} implemented an MDA training with deep networks.
For a detailed survey of MDA, readers are referred to Zhao et al. \cite{Zhao2020b}.
Zhao et al. \cite{Zhao2019a} extended Cycada \cite{Hoffman2018} to MDA for semantic segmentation.
A work of Nigam et al. \cite{Nigam2018}
utilizes multiple models trained with 
different source datasets as feature extractors, and adapt the segmentation layers to a target dataset
of overhead images from a drone.
Although their method is similar to ours in a way that multiple pre-trained networks are
exploited, their task is supervised domain adaptation where 
they prepare their own labeled dataset, while our method 
utilizes publicly available datasets that are not specifically built for greenhouse scenes.

\subsubsection{Benchmarks}
As a benchmark of semantic segmentation, a task considered de-facto standard is 
adaptation from synthetic datasets such as the GTA5 dataset \cite{Richter2016}
and the SYNTHIA dataset \cite{Ros2016} to a real dataset, most often Cityscapes dataset \cite{Cordts2016}.
In this task, scenes are limited to urban scenes in both the source datasets and the target datasets.
Besides, the set of object classes is identical in all the datasets.
In this paper, we further investigate the methodology of transferring knowledge
in cases where the structure of the environments is different,
and the source and the target datasets do not share their label space.

\section{Proposed method}
\label{proposed-method}
\subsection{Overview}
%
%
\begin{figure*}[tb]
  \centering\includegraphics[width=\textwidth]{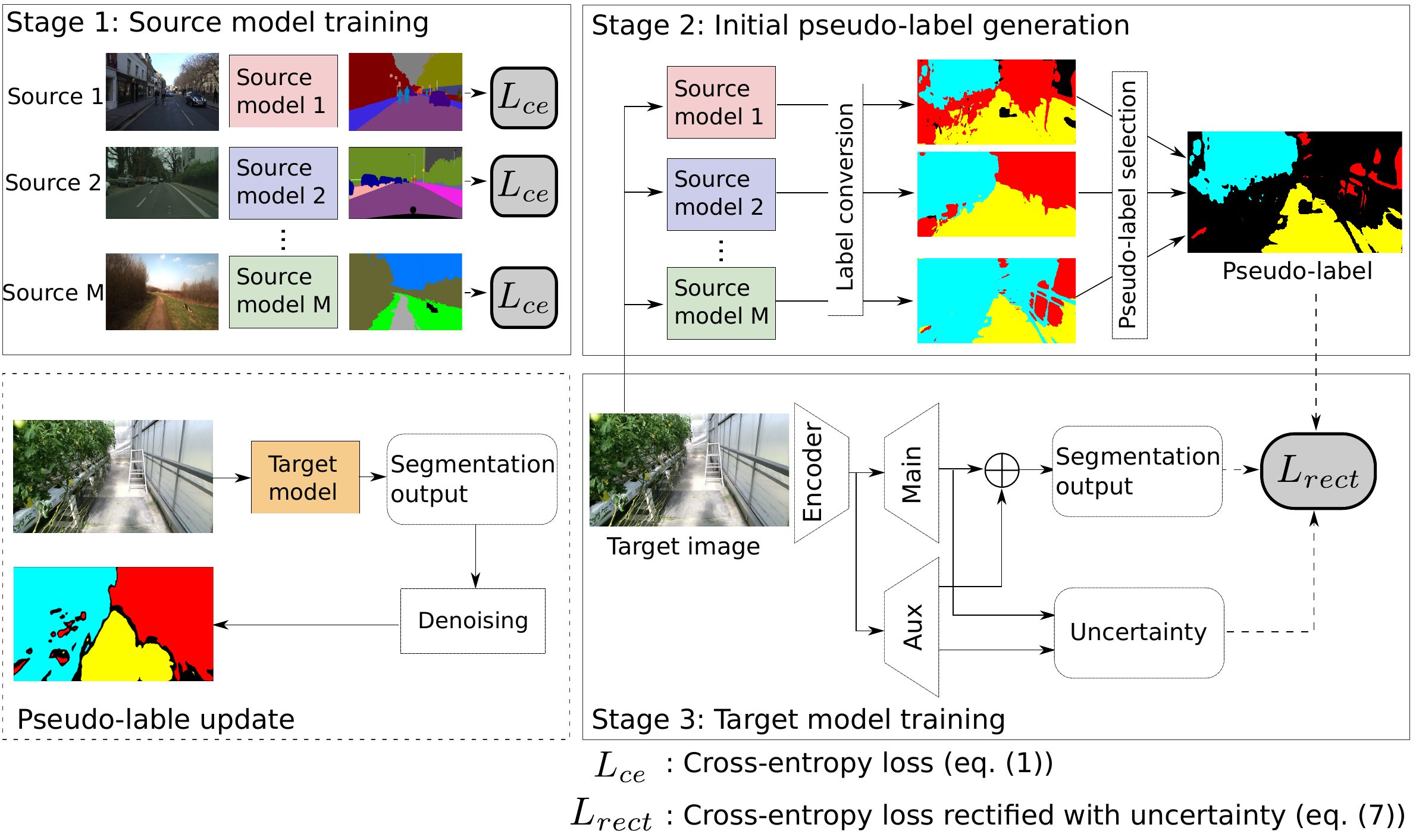}
  \caption{Overview of the proposed method.: Stage 1: Train models
    with the source datasets in supervised training.
    Stage 2: Pseudo-labels are generated for the
    target images using the output from the source models.
    Stage 3: Train a target model using the generated pseudo-labels.
    (Best viewed in color)}
  \label{fig:method_overview}
\end{figure*}
The purpose of this work is to train
a semantic segmentation model that achieves low error
on greenhouses images without hand-annotated labels.
To this end, we consider using multiple publicly available labeled image datasets
as source datasets to transfer knowledge to the target images.
Specifically, we use datasets of urban scenes and outdoor scenes,
namely, CamVid dataset \cite{Brostow2009}, Cityscapes dataset \cite{Cordts2016},
and Freiburg Forest dataset \cite{Valada2017}, as source datasets.
Each source dataset has a different set of object classes.
The target datasets are greenhouse images with the following classes:
\textit{plant}, \textit{artificial object}, and \textit{ground}.
The target datasets do not have any label data for supervision.

Formally, we assume $M$ source datasets $S_{1}, \dots, S_{M}$
and an unlabeled dataset $S_T$.
A source dataset $S_{i}$ consists of an image set $X_i=\{x_{i,j}\}^{N_i}_{j=1}$ and
a set of corresponding segmentation label maps $Y_i=\{y_{i,j}\}^{N_i}_{j=1}$,
where $N_i$ denotes the number of images in $S_i$.
The source datasets can have a different number of object classes.
The target dataset $S_T$ consists of only an image set $X_T=\{x_{T,j}\}^{N_T}_{j=1}$,
where $N_T$ denotes the number of images in $S_T$.



Fig. \ref{fig:method_overview} shows the overview of the proposed method.
It consists of three stages.
The first stage is the training of source models.
For each source dataset, a segmentation model is trained in a supervised manner.
In the second stage, the target images are fed into the source models
and pseudo-labels are generated by merging the outputs from the source models.
An output from each source model is first converted to the target labels
using a label conversion function, 
which is heuristically defined.
The pseudo-labels are then generated by selecting the labels that the models
unanimously predict as the same label.
Finally in the third stage, the target model is trained
with the pairs of the target images and corresponding pseudo-labels.
The classification loss is adaptively weighted by
the uncertainty of the prediction
as proposed in \cite{Zheng2021}.
During the training, the pseudo-labels are updated using
the outputs from the currently trained model for further leveraging the knowledge
learned from the initial pseudo-labels.
In the pseudo-label update, a denoising method inspired by \cite{Zhang2021} is used.

In our method, we employ ESPNetv2 \cite{Mehta2019}, a light-weight
semantic segmentation model, as base network architecture
for its suitability to real-time scene recognition of mobile robots.
The proposed method, however, is not dependent
on a specific network architecture.

\subsection{Model pre-training}

We first train a segmentation model on each source
in a fully supervised manner.
In this step, each model is trained with the original
classes in the corresponding source dataset.
A standard cross entropy loss is used as a loss function:
\begin{eqnarray}
  L_{ce}(x,y;\theta)=-\sum_{h, w}\sum_{c\in C}y^{\left(h,w,c\right)}
  \log F\left(x ; \theta\right)^{\left(h,w,c\right)},
\end{eqnarray}
where $F\left(x ; \theta\right)^{\left(h, w, c\right)}$ denotes the probability of
class $c$ at the location $(h, w)$ for an image $x$ predicted by
the segmentation model with weights $\theta$, and
$y^{\left(h,w\right)}$ denotes a one-hot label at pixel $(h,w)$.
A source model for dataset $S_i$ is initialized as follows:
\begin{eqnarray}
  \theta_i = \argmin_\theta \sum_{(x,y)\in S_i} L_{ce}(x,y;\theta).
\end{eqnarray}

Usually, in loss calculation,
greater weight is set for the loss of rarer classes
in order to deal with class imbalance.
For our task, however, we empirically found that
a model provides more accurate pseudo-labels
when the model is trained with a weight proportional to
the frequency of the label, which is contrary
to the usual practice.
We suppose the reason is as follows.
The usual loss weighting is mainly for capturing
relatively small objects such as road signs and poles.
In our target environment, i.e., greenhouses,
such a fine-grained perception is not necessary
because the space of the environment is fixed
and thus the variation of the scale of the objects is limited.
Moreover, the object classes in the target datasets are coarse.
Therefore, the usual practice of training may
lead to a model that is overly focused on small objects
which is unnecessary for the pseudo-label generation.

\subsection{Pseudo-label generation using multiple pre-trained models}
\label{sec:pseudo-label_generation}

\subsubsection{Underlying idea}

\begin{figure}[tb]
  \raggedleft
  \begin{minipage}{0.54\hsize}
    \centering\includegraphics[width=\hsize]{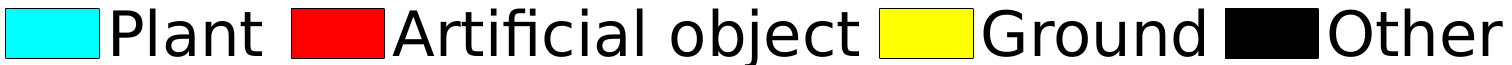}
  \end{minipage}\\

  \centering
  \begin{minipage}{0.32\hsize}
    \centering\includegraphics[width=\hsize]{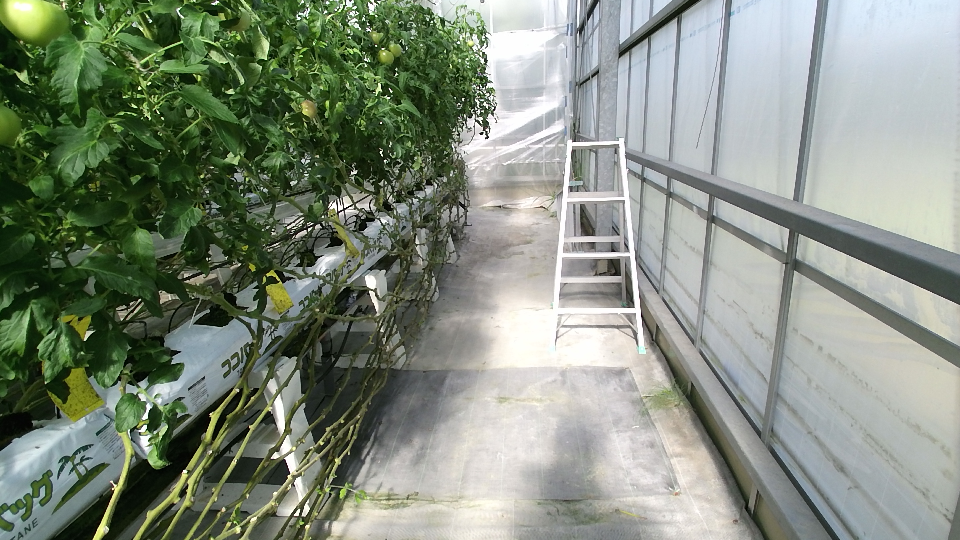}
    \subcaption{Camera image}
    \label{fig:pseudo-label_explain_camera}
 \end{minipage}
  \begin{minipage}{0.32\hsize}
    \centering\includegraphics[width=\hsize]{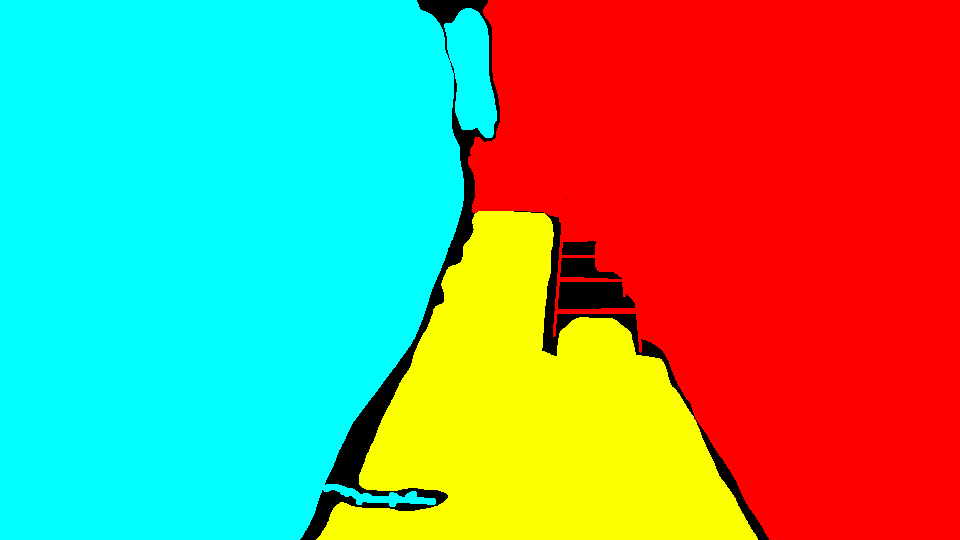}
    \subcaption{Ground truth}
    \label{fig:pseudo-label_explain_gt}
  \end{minipage}
  \begin{minipage}{0.32\hsize}
    \centering\includegraphics[width=\hsize]{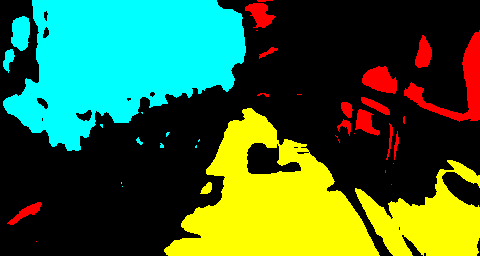}
    \subcaption{Pseudo-label}
    \label{fig:pseudo-label_explain_pseudo}
  \end{minipage} \\

  \begin{minipage}{0.32\hsize}
    \centering\includegraphics[width=\hsize]{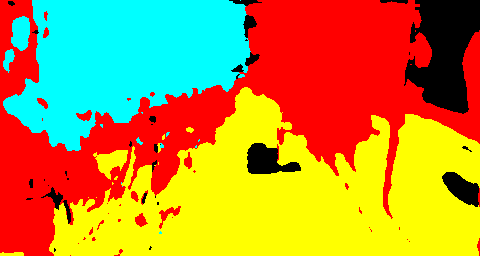}
    \subcaption{CamVid}
    \label{fig:pseudo-label_explain_camvid}
  \end{minipage}
  \begin{minipage}{0.32\hsize}
    \centering\includegraphics[width=\hsize]{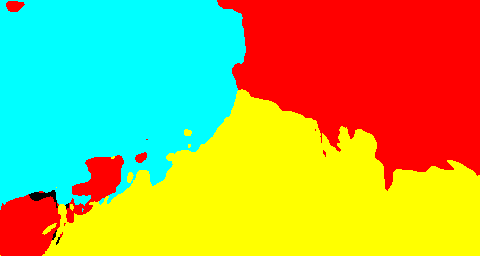}
    \subcaption{Cityscapes}
    \label{fig:pseudo-label_explain_city}
  \end{minipage}
  \begin{minipage}{0.32\hsize}
    \centering\includegraphics[width=\hsize]{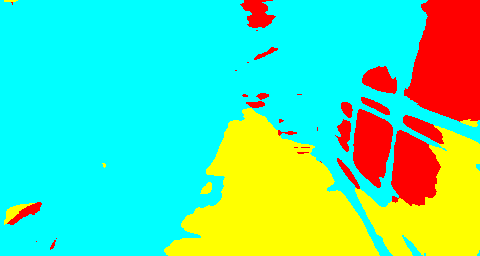}
    \subcaption{Freiburg Forest}
    \label{fig:pseudo-label_explain_forest}
  \end{minipage}
  \caption{Examples of output from each source model and resulting pseudo-label}
  \label{fig:pseudo-label_explain}
\end{figure}

Our intuition behind this method of multi-source pseudo-label generation
is that while the prediction from each source model
contains a lot of misclassifications,
the models correctly predict many of the pixels in common.
Fig. \ref{fig:pseudo-label_explain} shows outputs for a target image
from different source models 
trained with the CamVid dataset \cite{Brostow2009},
Cityscapes dataset \cite{Cordts2016} and the Freiburg Forest dataset \cite{Valada2017}
as well as the target image, hand-labeled ground truth, and the pseudo-label
generated from the three outputs.
As can be seen in Fig. \ref{fig:pseudo-label_explain},
many of the true regions of ``plant'' class and ``ground'' class are
correctly classified in all the models.
This implies that 
getting a consensus of the models is effective for generating precise pseudo-labels 
as shown in Fig. \ref{fig:pseudo-label_explain_pseudo}.
Another thing worth noting is that each model has its own characteristics of prediction.
For example, 
the prediction of the model trained with Freiburg Forest dataset on the ground regions
better captures the shape of true ground regions than the other two models.
However, the prediction on the plants and artificial objects is not accurate.
On the other hand, in the model trained with Cityscapes,
the prediction on the ground region includes a lot of false positives, while
that on the other objects is more accurate than the Freiburg Forest model.
These differences stem from the structural differences of the source datasets. 
The proposed method avoids biased training to the structural features of
a specific source dataset
dissimilar to the target by getting agreements of multiple source models.

\subsubsection{Pseudo-label generation method}

In this step, pseudo-labels for the target images
are generated using outputs from the models
pre-trained with the source datasets.
First, the target images are fed in all the pre-trained
models to yield predicted labels in the label space
of each source domain.
The predicted label from the model of source $i$ for the $j$th target image
is given as follows:
\begin{equation}
  p_{i,j}^{(h,w)}=\argmax_c F\left(x_{T,j}; \theta_i\right)^{\left(h,w,c\right)}.
\end{equation}
The predicted labels 
are then mapped to the target label space
using a label conversion function
$\xi_i:\mathcal{Y}_i\rightarrow \mathcal{Y}_T$,
where $\mathcal{Y}_i$ and $\mathcal{Y}_T$
denote a set of labels in source $S_i$ and target $S_T$, respectively:
\begin{equation}
  {p^{\prime}_{i,j}}^{(h,w)}=\xi_i(p_{i,j}^{(h,w)}).
\end{equation}
Here, we heuristically define the mapping from
the source classes to the target classes
for each source domain as shown in Table \ref{table:label_mapping}.
After the label conversion, for each pixel of the target images,
a predicted label is assigned to the pixel
if the prediction is the same among all the source models.
Otherwise, ``other'' label is assigned and not used for training.
Formally, a pseudo-label for the pixel $(h, w)$
of the $j$th target image $x_{T,j}$ is generated as follows:
\begin{align}
  \hat{y}_{j}^{(h,w)} =\psi \left({p^{\prime}_{1,j}}^{(h,w)},\cdots,{p^{\prime}_{M,j}}^{(h,w)}\right)=
  \begin{cases}
    c_k      & \text{if $\forall i \in \{1,...,M\}, p^{\prime}_{i,j}=c_k$} \\
    c_{\phi} & \text{otherwise,}
  \end{cases}
\end{align}
where $c_k\in \mathcal{Y}_T$ is a class category in the target dataset
and $c_{\phi}$ is a class label ``other'' that is 
not used for loss calculation during the training
and thus does not affect the training.

\subsection{Model training on the target data}
\label{sec:model_training}

We train a segmentation model
with the target images and the corresponding pseudo-labels.
Although the valid labels in the pseudo-labels generated
in Sec. \ref{sec:pseudo-label_generation}
are unanimously predicted by the source models and thus highly likely correct,
we introduce uncertainty-based loss weighting and
prototype-based pseudo-label denoising for further
suppressing the effect of noise in the pseudo-labels.

\subsubsection{Loss weighting based on the uncertainty of estimation}
\label{sec:model_training_loss_weighting}

In standard training of semantic segmentation models,
the cross entropy loss is widely used.
In the training with pseudo-labels, however,
equally treating all the pseudo-labels may lead to poor performance
due to noise in the pseudo-labels.
Zheng et al. \cite{Zheng2021} proposed 
an adaptive loss weighting that decays as the uncertainty
of the estimation increases
so that the losses on the pixels with high uncertainty
do not affect the training much.
Their method exploits an auxiliary segmentation branch
to predict pixel-wise uncertainty of the label prediction.
We attach an auxiliary segmentation branch to ESPNetv2 for uncertainty estimation.
The entire segmentation network is shown in Fig. \ref{fig:network_detail}
in Appendix \ref{sec:network_architecture}.

Given the output of primary branch $F(x;\theta)$
and that of the auxiliary branch $F_{aux}(x;\theta)$,
the uncertainty of the prediction is defined as
Kullback-Leibler Divergence (KLD) of the distributions
of class probabilities from the two branches:
\begin{eqnarray}
  \label{eq:kld}
  D_{kl}\left(x; \theta \right)^{\left(h, w\right)} = \sum_{c\in C}F(x;\theta)^{\left(h, w, c\right)}
  \log{\frac{F(x;\theta)^{\left(h, w, c\right)}}{F_{aux}(x;\theta)^{\left(h, w, c\right)}}}.
\end{eqnarray}
Using eq. \eqref{eq:kld}, 
the loss function is defined as follows:
\begin{eqnarray}
  L_{rect}\left(x, \hat{y}; \theta \right)=\sum_{h,w}\left(\exp{\{-D_{kl}\left(x; \theta \right)^{\left(h, w\right)}\}}
  L_{ce}\left(x, \hat{y}; \theta\right)^{\left(h, w\right)}+D_{kl}\left(x; \theta \right)^{\left(h, w\right)}\right).
\end{eqnarray}

\subsubsection{Prototype-based pseudo-label denoising}
\label{sec:model_training_prototype_denoising}

In the middle of the training,
we update the pseudo-labels with the outputs from the current model
to further leverage the knowledge learned from the initial pseudo-labels.
We introduce pseudo-label updating with prototype-based denoising
inspired by Zhang et al. \cite{Zhang2021}.

At first, the prototypes are calculated.
The prototype of class $c$ is a mean of the features that belong to the class $c$
and is calculated using the current pseudo-labels as follows:
\begin{equation}
  \eta^{(c)}=\frac{\sum_{(x,\hat{y}) \in S_{T}} \sum_{h,w} f^{(h,w)} * \mathbbm{1}\left(\hat{y}^{(h,w)}==c\right)}{\sum_{(x,\hat{y}) \in S_{T}} \sum_{h,w} \mathbbm{1}\left(\hat{y}^{(h,w)}==c\right)},
  \label{eq:initial_prototype}
\end{equation}
where $f^{(h,w)}$ is an intermediate feature provided by the currently trained model
$F\left(\cdot ; \theta\right)$
at pixel $(h, w)$, and $\mathbbm{1}$ is the indicator function
that returns 1 when its argument is true and 0 otherwise.

Pseudo-labels for image $x_{T,j}$ are then updated as follows:
\begin{equation}
  {\hat{y}_j}^{(h,w)}=
  \begin{cases}
    \argmax_c \left(\omega^{(h,w,c)}F\left(x_{T,j}; \theta\right)^{\left(h,w,c\right)}\right)
             & \text{if $\max_c \left(\frac{\omega^{(h,w,c)}p_{j}^{\left(h,w,c\right)}}{\mu}\right) > \alpha$} \\
    c_{\phi} & \text{otherwise,}
  \end{cases}
  \label{eq:pseudo_label_update}
\end{equation}
where $\mu=\sum_{c^{\prime}} \omega^{(h,w,c^{\prime})}p_{j}^{\left(h,w,c^{\prime}\right)}$
is a normalization factor and
$\alpha$ is a threshold for pseudo-label selection,
which is set to 0.9 in the experiments.
$\omega$ is a vector of weight values to modulate the class probabilities,
calculated using the distances between the feature and each prototype.
For each pixel, the weight $\omega$ is calculated as follows:
\begin{equation}
  \omega^{(h,w,c)}=\frac{\exp \left(-\left\|{f}^{(h,w)}-\eta^{(c)}\right\| / \tau\right)}{\sum_{c^{\prime}} \exp \left(-\left\|{f}^{(h,w)}-\eta^{\left(c^{\prime}\right)}\right\| / \tau\right)},
\end{equation}
where $\tau$ is a temperature parameter that controls
the degree of bias of the distribution, which is set to 1 in \cite{Zhang2021}.
The values of $\omega$ indicate the class likelihood of the feature based on 
the distance from the feature to each prototype,
and the probabilities predicted by $F\left(\cdot ; \theta\right)$
are rectified by $\omega$.

\subsection{The overall algorithm}
\label{sec:overall_algorithm}

The overall algorithm is shown in Algorithm \ref{alg1}.
As already mentioned, our training pipeline consists of three stages.
A segmentation model for each source dataset is trained
in standard supervised learning in the first stage
and initial pseudo-labels for the target images are generated
using the pre-trained models in the second stage.
In the third stage, the target model is trained using the target images
and the pseudo-labels.
The training is done in several ``rounds'', each of which consists of
several epochs.
Here, the number of epochs per round is set to 5, 
and the maximum number of rounds is set to 10.

During training, the pseudo-labels are updated using the target model
currently being trained.
In this work, we update the pseudo-labels once after
the third round of training with the initial pseudo-labels
because the training with the initial pseudo-labels converges by
the third round.
We compare the different pseudo-label update strategies
in Sec. \ref{sec:result-pseudo-label-update-strategy}.

\begin{algorithm}[tb]
  \caption{Multi-source pseudo-label learning}
  \label{alg1}
  \begin{algorithmic}[1]
    \REQUIRE Source dataset $S_i$, label conversion functions $\xi_i$ with $i\in \{1,\cdots, M\}$,
    Target dataset $S_T=\{X_T\}$
    \ENSURE The target model $F(\cdot;\theta_T)$
    \STATE \# Step 1: Source model pre-training
    \STATE Train the source model $F(\cdot;\theta_i)$ with $S_i=\{X_i, Y_i\}$ and $L_{ce}$
    \STATE
    \STATE \# Step 2: Generating initial pseudo-labels
    \STATE Get outputs for the target images: $p_i \leftarrow F\left(x_T;\theta_i\right)$
    \STATE Label conversion: $p^{\prime}_i \leftarrow \xi_i\left(p_i\right)$
    \STATE Generate pseudo-labels: $\hat{y} \leftarrow \psi\left(p^{\prime}_1, \cdots, p^{\prime}_M\right)$
    \STATE
    \STATE \# Step 3: Target model training
    \FOR{\text{$round \leftarrow 0$ to $max\_round$}}
    \IF{$is\_round\_to\_update\_pseudo\_labels$}
    \STATE Calculate the prototypes by eq. (\ref{eq:initial_prototype})
    \STATE Update the pseudo-labels by eq. (\ref{eq:pseudo_label_update})
    \ENDIF
    \STATE
    \FOR{\text{$k \leftarrow 0$ to $epoch\_per\_round$}}
    \FOR{\text{$j \leftarrow 0$ to $len(X_T)$}}
    \STATE Train the model with $\{X_T, \hat{Y}\}$ and $L_{rect}$
    \ENDFOR
    \ENDFOR
    \ENDFOR
  \end{algorithmic}
\end{algorithm}

\hypertarget{experiments}{%
  \section{Experiments}\label{experiments}}

\begin{table*}[t]
  \centering
  \caption{Label conversion from the source datasets to the target sets}
  \label{table:label_mapping}
  \small
  \begin{tabular}{cccc}
    \toprule
    \textbf{CamVid} & \textbf{Cityscapes} & \textbf{Forest}                  & \textbf{Greenhouse A, B, C} \\
    \cmidrule(lr){1-3} \cmidrule(lr){4-4}
    Tree            & Vegetation          &
    \begin{tabular}{c}
      Grass, Tree
    \end{tabular}
                    & Plant                                                                                \\
    \cmidrule(lr){1-3} \cmidrule(lr){4-4}
    \begin{tabular}{c}
      Building, Pole, \\ SignSymbol, Fence, \\Car, Road\_marking
    \end{tabular}
                    &
    \begin{tabular}{c}
      Building, Wall \\Fence, Pole, \\Traffic light, \\Traffic sign, \\Car, Truck,
      \\Bus, Train \\Motorcycle, Bicycle
    \end{tabular}
                    & Obstacle            & Artificial object                                              \\
    \cmidrule(lr){1-3} \cmidrule(lr){4-4}
    \begin{tabular}{c}
      Road, Pavement
    \end{tabular}
                    &
    \begin{tabular}{c}
      Road, Sidewalk, \\Terrain
    \end{tabular}
                    & Road                & Ground                                                         \\
    \cmidrule(lr){1-3} \cmidrule(lr){4-4}
    \begin{tabular}{c}
      Sky, Pedestrian \\ Bicyclist, Unlabeled
    \end{tabular}
                    &
    \begin{tabular}{c}
      Sky, Person \\ Rider, Background
    \end{tabular}
                    & Sky                 & Other (Not used in the training)                               \\
    \bottomrule
  \end{tabular}
\end{table*}

\subsection{Experimental setup}

\subsubsection{Training conditions}

We use PyTorch implementation of ESPNetv2 \cite{Mehta2019}.
For estimating uncertainty, an auxiliary segmentation branch is
attached. 
The architecture is shown in
Fig. \ref{fig:network_detail} in Appendix \ref{sec:network_architecture}.
The processing time was about 44 [msec/image], or 23 [fps].
The number of floating point operations (FLOPs) is 0.79 [GFLOPs]
while that of the original ESPNetv2 for semantic segmentation 
is 0.76 [GFLOPs] \cite{Mehta2019}, indicating that 
attatching the auxiliary branch did not affect
the computational efficiency.
All training of the proposed method is performed on one NVIDIA Quadro RTX 8000
with 48GB RAM.
The numbers of training rounds and epochs per round
are set to 10 and 5, respectively.
The learning rate is fixed to $5\times 10^{-4}$
and the batch size is 48.
Adam \cite{Kingma2015} is used as an optimizer.
Weights of the target models are initialized by the supervised training of
semantic segmentation on the CamVid dataset \cite{Brostow2009},
except for the last classification layer initialized with random values.
We measure segmentation performance with the Intersection over Union (IoU) metric.

\subsubsection{Datasets}

We use CamVid \cite{Brostow2009}, Cityscapes \cite{Cordts2016} and
Freiburg Forest dataset \cite{Valada2017}
(hereafter referred to as CV, CS, and FR, respectively)
for the training of the source models.
CV, CS, and FR have 367, 2975, and 230 labeled training images, respectively.
Label conversions from the source datasets to the target datasets
($\xi_i$) are heuristically defined as shown in Table \ref{table:label_mapping} .
A more detailed analysis of the label mapping is in Appendix \ref{sec:appendix_pred_dist}.

\begin{figure}[tb]
  \centering
  \begin{minipage}{0.26\hsize}
    \centering
    \includegraphics[width=\hsize]{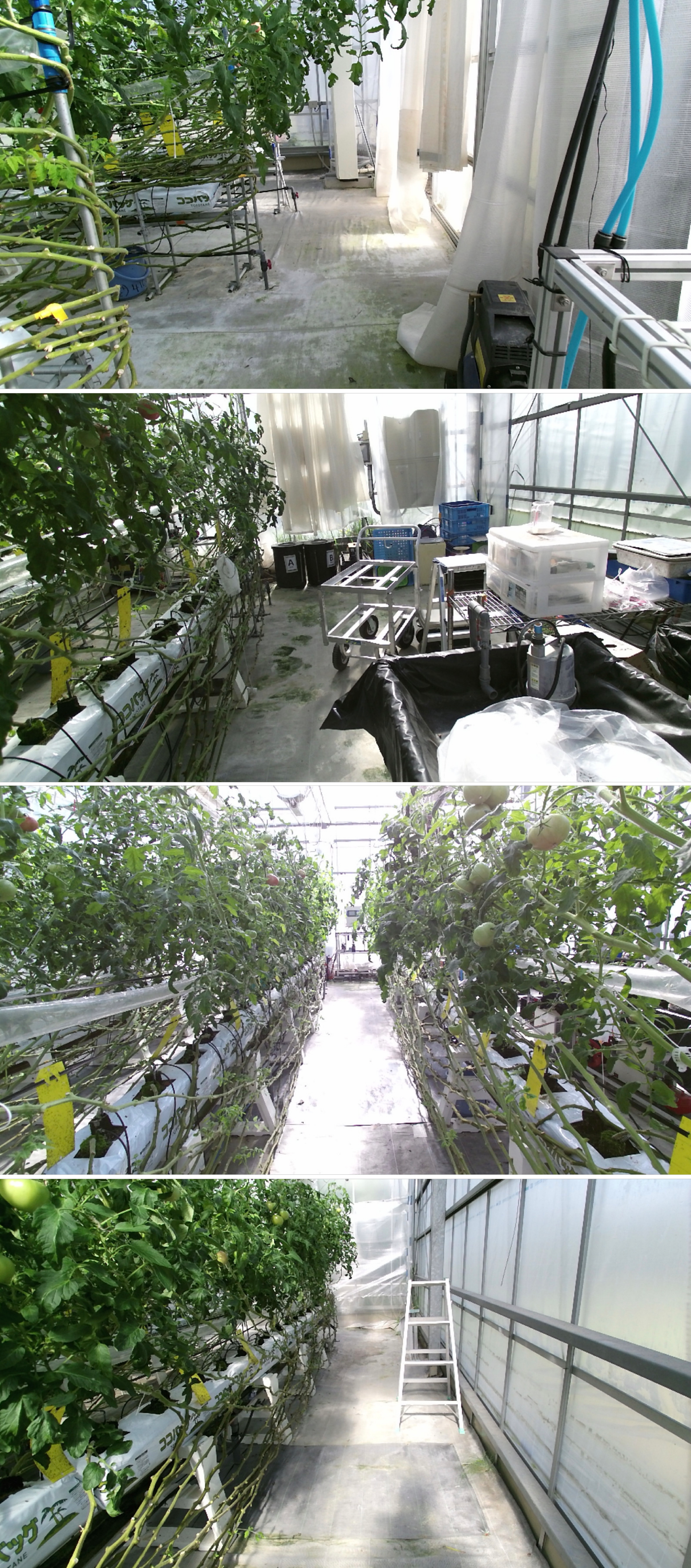}
    \subcaption{Greenhouse A}
    \label{fig:dataset_example_greenhouse1}
  \end{minipage}
  \begin{minipage}{0.26\hsize}
    \centering
    \includegraphics[width=\hsize]{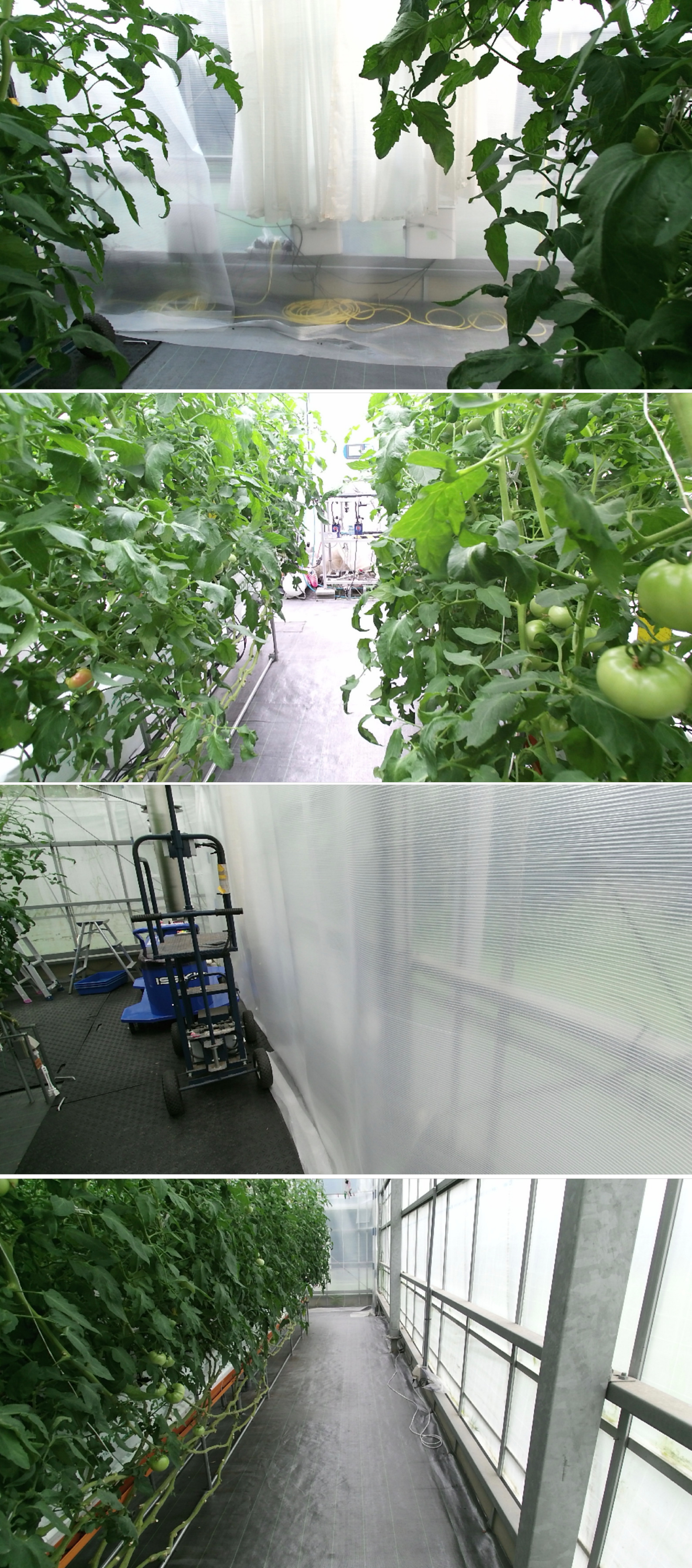}
    \subcaption{Greenhouse B}
    \label{fig:dataset_example_greenhouse2}
  \end{minipage}
  \begin{minipage}{0.26\hsize}
    \centering
    \includegraphics[width=\hsize]{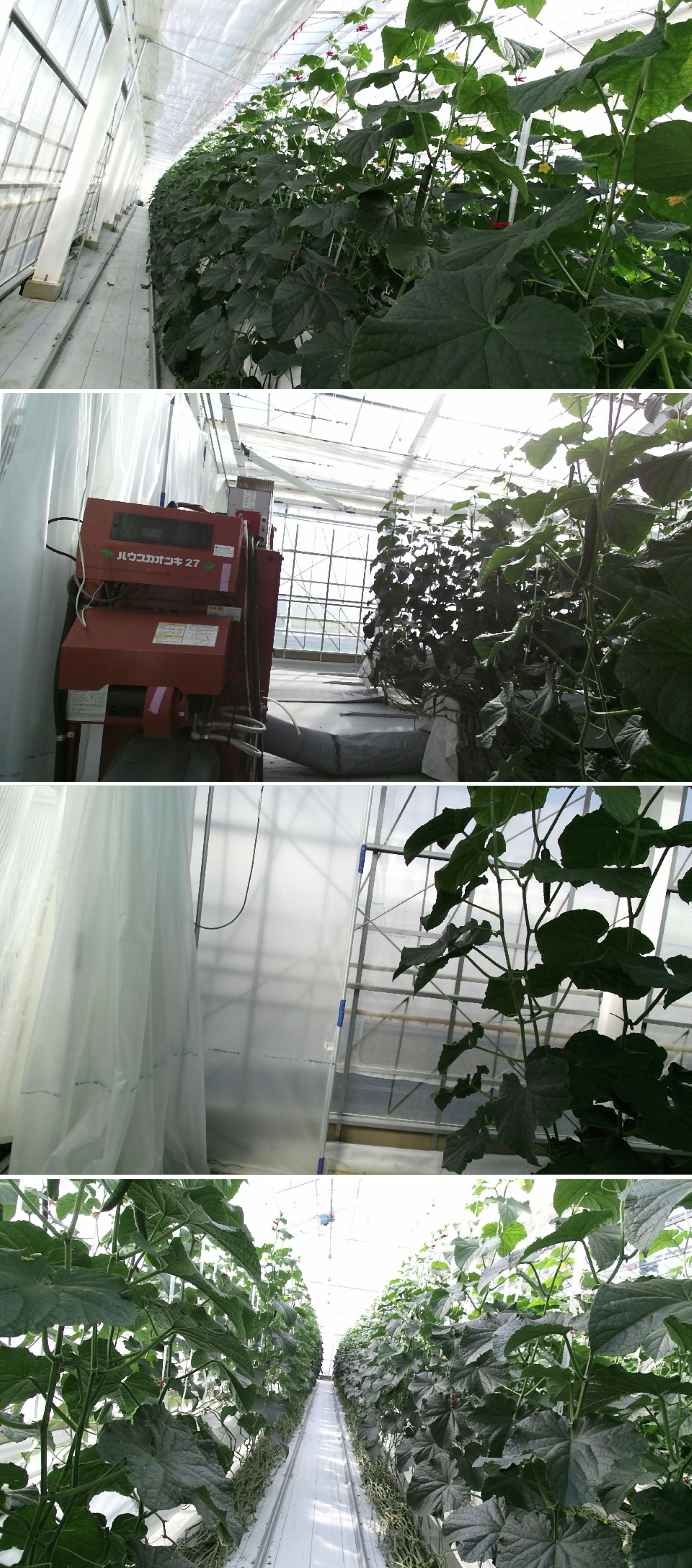}
    \subcaption{Greenhouse C}
    \label{fig:dataset_example_cucumber}
  \end{minipage}
  \caption{Example of images of the target datasets. Greenhouse A and B are
    the same greenhouse that grows tomatoes, but taken in different times and thus the lighting condition
    and growth of leaves differ. Greenhouse C is a greenhouse of cucumber that has
    different appearance from tomato plants.}
  \label{fig:dataset_example}
\end{figure}

As target greenhouse datasets for adaptation, we acquired images from three
greenhouses; Greenhouse A that grows tomatoes,
Greenhouse B that is the same
greenhouse as Greenhouse A but acquired in different date and time, and
Greenhouse C that grows cucumbers.
Each target dataset has 6689 unlabeled images.
Example images of the target datasets are shown in Fig. \ref{fig:dataset_example}.
\begin{figure}[tb]
  \raggedleft
  \begin{minipage}{0.45\hsize}
    \centering\includegraphics[width=\hsize]{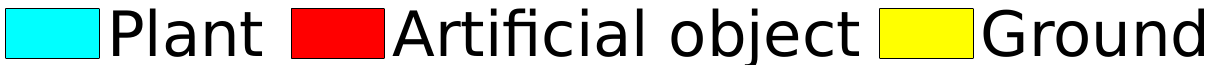}
  \end{minipage}\\

  \centering
  \begin{minipage}{0.26\hsize}
    \centering
    \includegraphics[width=\hsize]{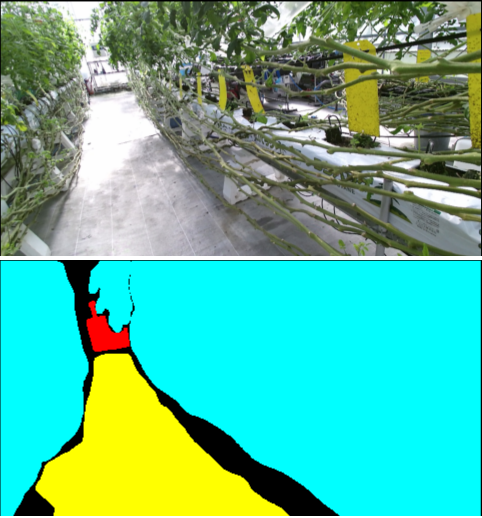}
    \subcaption{Greenhouse A}
    \label{fig:test_dataset_example_greenhouse1}
  \end{minipage}
  \begin{minipage}{0.26\hsize}
    \centering
    \includegraphics[width=\hsize]{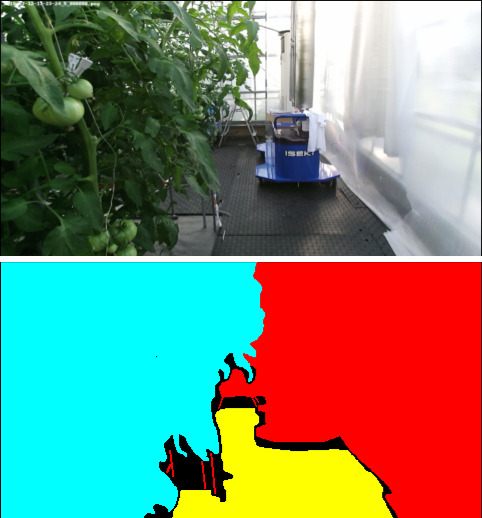}
    \subcaption{Greenhouse B}
    \label{fig:test_dataset_example_greenhouse2}
  \end{minipage}
  \begin{minipage}{0.26\hsize}
    \centering
    \includegraphics[width=\hsize]{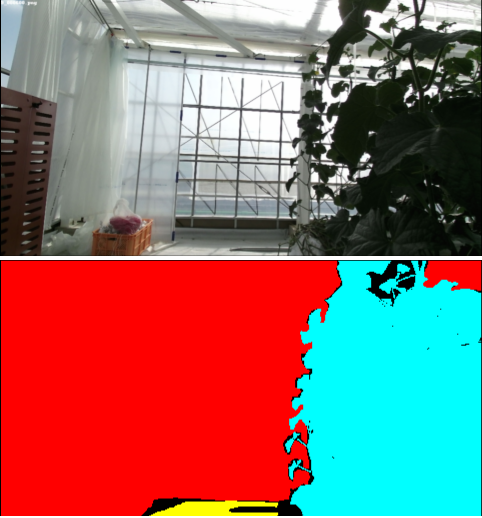}
    \subcaption{Greenhouse C}
    \label{fig:test_dataset_example_cucumber}
  \end{minipage}
  \caption{Example of the test images.
    Regions of each object class are roughly labeled, rather than labeled with pixel-level precision.
    For example, the plant rows in Greenhouse A are uniformly labeled as ``plant''}
  \label{fig:test_dataset_example}
\end{figure}
For each target dataset,
we prepared test images with manually annotated labels for evaluation.
The numbers of the test images are 100 for Greenhouse A,
and 50 for B and C.
The labels on the test images include the ``other'' class and
the pixels with the class are not considered in the calculation of the metric
(See Fig. \ref{fig:test_dataset_example}).
More detailed annotation policy is described in Appendix \ref{sec:label_policy}.

\hypertarget{Results}{%
  \subsection{Comparison to single-source baselines}\label{results}}

\begin{table*}[tb]
  \centering
  \caption{Result of the adaptation in mean IoU
    (CV: CamVid, CS: Cityscapes, FR: Freiburg Forest.
    The best result for each target is shown in \textbf{bold}
    and the second best result is \underline{underlined}.)}
  \label{table:experiment1}
  \begin{tabular}{ c  c  c  c  c  c }
    \toprule
    \multirow{2}[0]{*}{\textbf{Target}} & \multirow{2}[0]{*}{\textbf{Source}}
                                        & \multicolumn{3}{c}{\textbf{Class IoU}} & \multirow{2}[0]{*}{\textbf{mIoU}}                                                                      \\
    \cmidrule{3-5}
                                        &                                        & \textbf{Plant}                    & \textbf{Artificial object} & \textbf{Ground}   &                   \\
    \midrule
    \multirow{10}[0]{*}{Greenhouse A}
                                        & CV (no adapt)                          & 64.72                             & 56.19                      & 46.34             & 55.75             \\
                                        & CS (no adapt)                          & 63.71                             & 68.71                      & 36.37             & 56.26             \\
                                        & FR (no adapt)                          & 60.24                             & 45.27                      & 30.49             & 45.33             \\
    \cmidrule[0.03em](lr){2-6}
                                        & CV                                     & 70.21                             & 68.66                      & 57.10             & 65.32             \\
                                        & CS                                     & 60.17                             & 71.47                      & 31.02             & 54.22             \\
                                        & FR                                     & 52.73                             & 50.68                      & 40.84             & 48.11             \\
                                        & CV+CS                                  & 73.81                             & 73.49                      & 47.55             & 64.94             \\
                                        & CS+FR                                  & \underline{79.50}                 & \underline{76.35}          & \underline{70.32} & \underline{75.39} \\
                                        & FR+CV                                  & 77.40                             & 73.00                      & 63.20             & 71.20             \\
                                        & CV+CS+FR                               & \textbf{80.71}                    & \textbf{78.21}             & \textbf{72.68}    & \textbf{77.20}    \\
    \midrule
    \multirow{10}[0]{*}{Greenhouse B}
                                        & CV (no adapt)                          & 65.57                             & 65.30                      & 57.65             & 62.84             \\
                                        & CS (no adapt)                          & 83.33                             & 80.09                      & 49.68             & 71.03             \\
                                        & FR (no adapt)                          & 48.47                             & 54.34                      & 19.17             & 40.66             \\
    \cmidrule[0.03em](lr){2-6}
                                        & CV                                     & 71.37                             & 77.15                      & 62.73             & 70.42             \\
                                        & CS                                     & 83.49                             & 77.39                      & 44.53             & 68.47             \\
                                        & FR                                     & 47.04                             & 58.60                      & 42.87             & 49.50             \\
                                        & CV+CS                                  & \textbf{87.19}                    & 87.00                      & 60.96             & 78.38             \\
                                        & CS+FR                                  & 82.56                             & \textbf{89.44}             & \underline{70.60} & \underline{80.87} \\
                                        & FR+CV                                  & 75.99                             & 85.70                      & 60.48             & 74.06             \\
                                        & CV+CS+FR                               & \underline{83.56}                 & \underline{89.08}          & \textbf{73.37}    & \textbf{82.33}    \\
    \midrule
    \multirow{10}[0]{*}{Greenhouse C}
                                        & CV (no adapt)                          & 58.89                             & 58.29                      & 19.82             & 45.66             \\
                                        & CS (no adapt)                          & 79.09                             & 81.10                      & 18.11             & 59.43             \\
                                        & FR (no adapt)                          & 61.65                             & 49.24                      & 25.56             & 45.48             \\
    \cmidrule[0.03em](lr){2-6}
                                        & CV                                     & 69.96                             & 70.30                      & 22.52             & 54.26             \\
                                        & CS                                     & 85.51                             & 85.30                      & 21.86             & 64.23             \\
                                        & FR                                     & 66.00                             & 65.52                      & 0.6549            & 44.06             \\
                                        & CV+CS                                  & 88.76                             & 87.20                      & 28.61             & 68.19             \\
                                        & CS+FR                                  & 84.99                             & 82.26                      & 29.56             & 65.61             \\
                                        & FR+CV                                  & \underline{91.12}                 & \textbf{90.00}             & \textbf{45.12}    & \textbf{75.41}    \\
                                        & CV+CS+FR                               & \textbf{91.16}                    & \underline{88.84}          & \underline{40.67} & \underline{73.56} \\
    \bottomrule
  \end{tabular}
\end{table*}

We conducted an experiment of adaptation from the source outdoor datasets to
each of the target greenhouse datasets.
In the single source baselines (CV, CS, and FR),
the initial pseudo-labels were generated via
the pseudo-label update procedure described
in Sec. \ref{sec:model_training_prototype_denoising}
in the source label space $\mathcal{Y_i}$, followed by
the label conversion by $\xi_i$ to map the labels to
the target label space.
All the models were trained with the initial pseudo-labels
for the first three rounds.
After that, the pseudo-labels were updated
using the hard pseudo-labels with the prototype-based denoising
and used in the rest of the training.

Table \ref{table:experiment1} shows the result of the training.
The results of ``no adapt'' were provided by directly feeding the target images to
the source models followed by the label conversion.
Before the adaptation, the IoUs of the source models
were low and not applicable to real applications.
The performance of single-source training
depended on the source dataset.
In some cases, the adaptation resulted in improvement from the model without adaptation,
while in some others the performance deteriorated after the adaptation.
The cause of the degradation is that the training progressed in the wrong direction
due to the noisy pseudo-labels. 
This indicates that
the training with a single source may not be reliable enough
when the source and target datasets have a large domain shift.

With our multi-source adaptation
, on the other hand, 
the accuracy was significantly improved
even though each source model is not reliable on the target data.
The results show the ability of our method to transfer useful common knowledge
about the appearance of objects from the source datasets to the target. 
The two-source training resulted in comparative or
even better performance than the three-source training (CV+CS+FR).
It is worth noting that
the combination of different types of environments results in
especially high accuracy, i.e., FR+CV and CS+FR.
CV (CamVid) and CS (Cityscapes) are datasets of urban scenes,
while FR (Freiburg Forest) consists of images of outdoor scenes
with rough terrains and vegetation,
This implies that 
we should use datasets with a high variety
of environments as sources.

The most suitable combination, however, depends on
the target dataset in the two-source setting.
For example, CS+FR resulted in the second-best performance
on Greenhouse A and B, while it was the worst
among the multi-source settings on Greenhouse C.
On the other hand, the three-source training
consistently performed the best or the second-best
on all the target datasets.
We, therefore, suggest exploiting the three source datasets
for the segmentation task in greenhouse environments.

\begin{figure*}[tb]
  \raggedleft
  \begin{minipage}{0.45\hsize}
    \centering\includegraphics[width=\hsize]{legend.pdf}
  \end{minipage}\\

  \centering

  \begin{minipage}{0.16\hsize}
    \centering\includegraphics[width=\hsize]{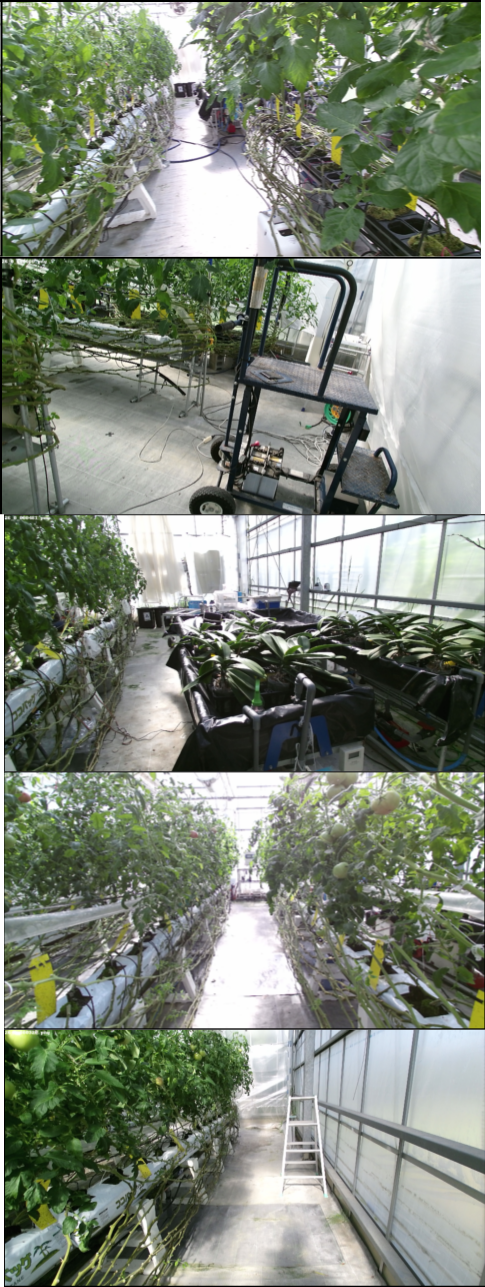}
    \subcaption{Camra image}
    \label{fig:greenhouse_a_camera}
  \end{minipage}
  \begin{minipage}{0.16\hsize}
    \centering\includegraphics[width=\hsize]{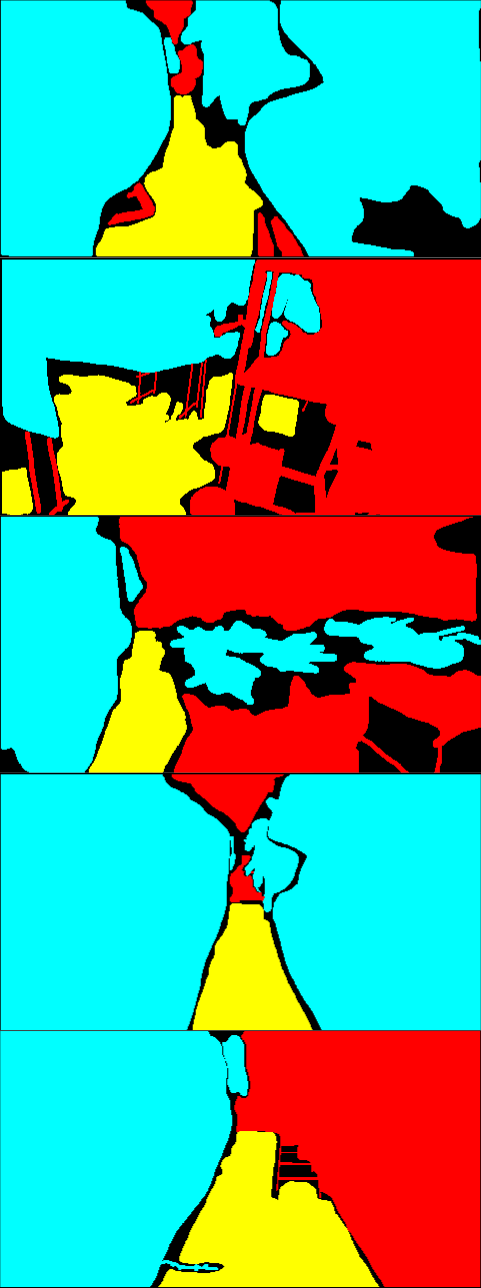}
    \subcaption{Ground truth}
    \label{fig:greenhouse_a_gt}
  \end{minipage}
  \begin{minipage}{0.16\hsize}
    \centering\includegraphics[width=\hsize]{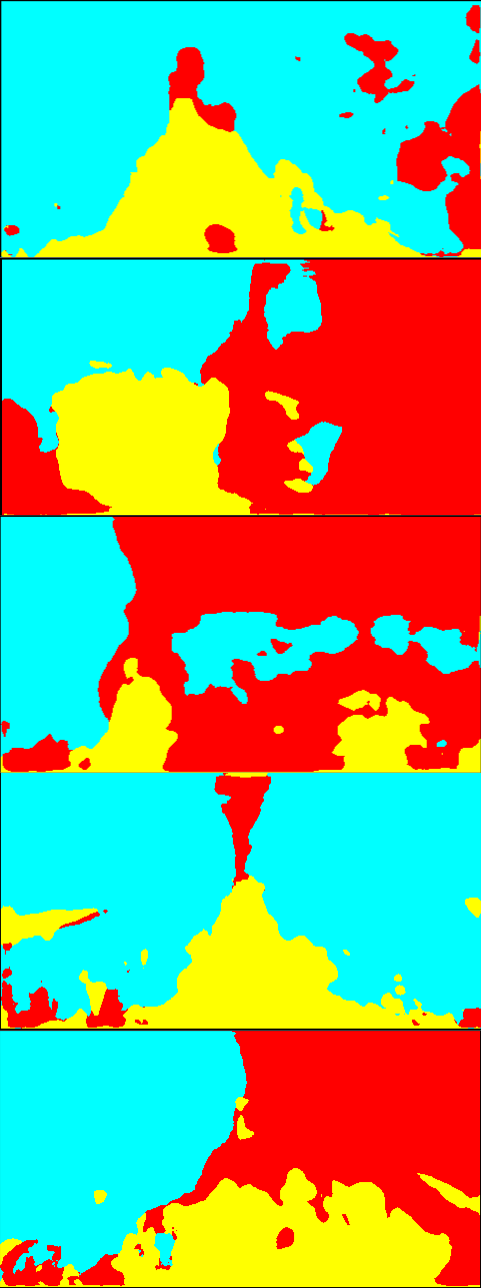}
    \subcaption{CV}
    \label{fig:greenhouse_a_cv}
  \end{minipage}
  \begin{minipage}{0.16\hsize}
    \centering\includegraphics[width=\hsize]{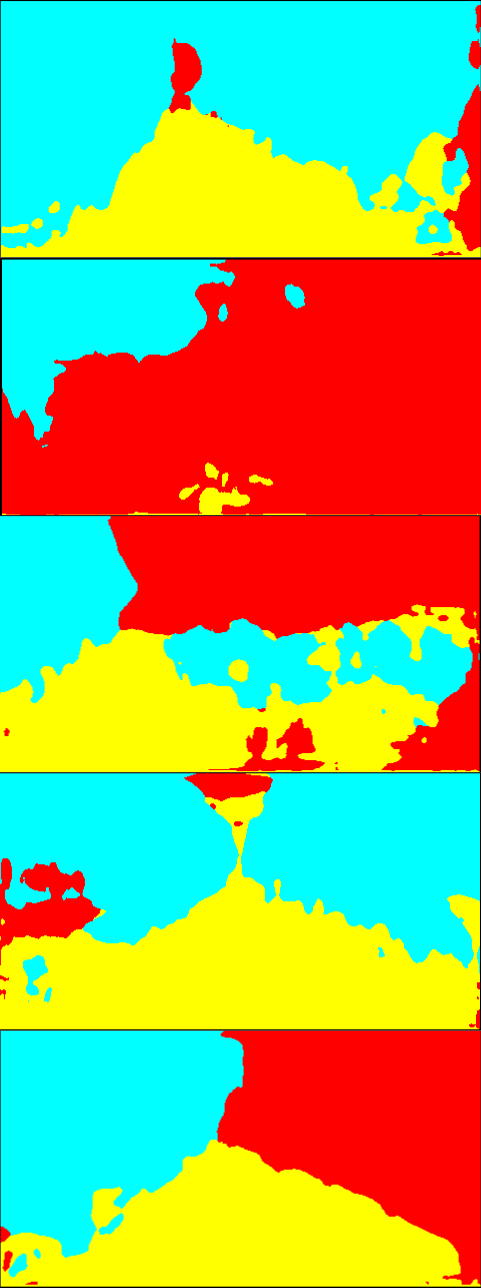}
    \subcaption{CS}
    \label{fig:greenhouse_a_cs}
  \end{minipage}
  \begin{minipage}{0.16\hsize}
    \centering\includegraphics[width=\hsize]{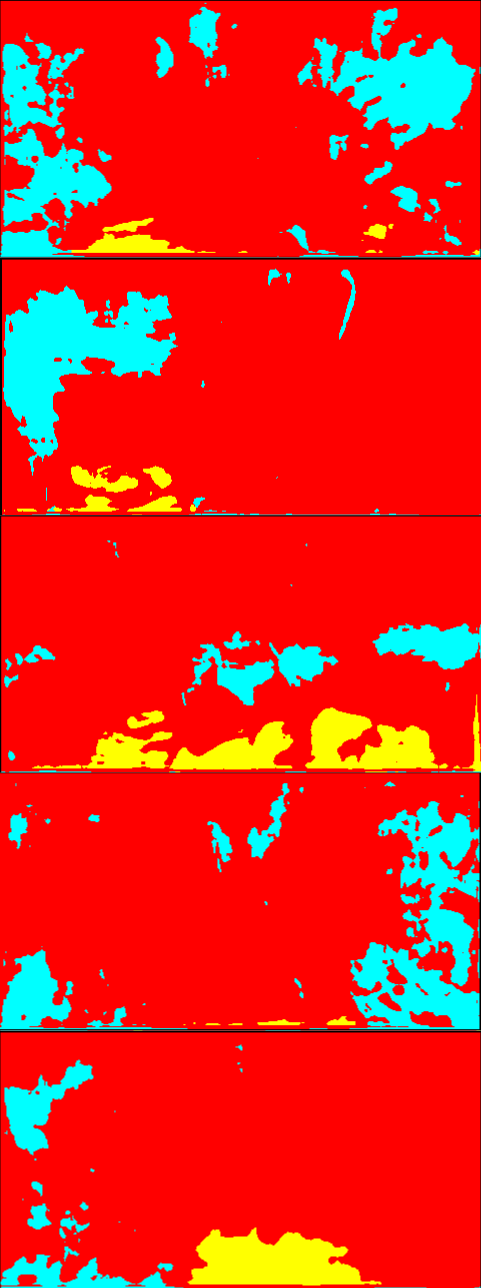}
    \subcaption{FR}
    \label{fig:greenhouse_a_fr}
  \end{minipage}
  \begin{minipage}{0.16\hsize}
    \centering\includegraphics[width=\hsize]{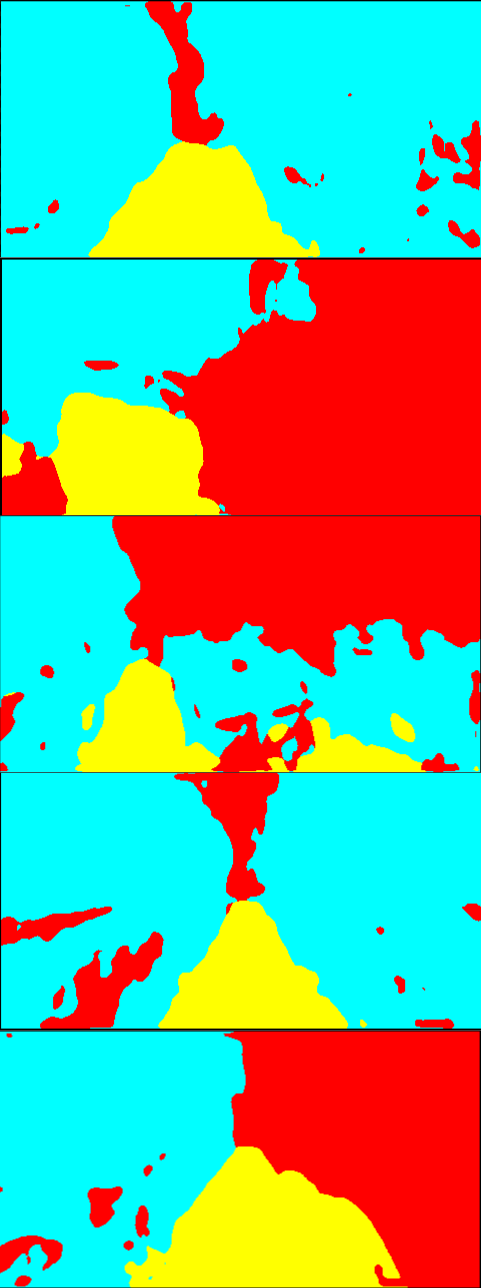}
    \subcaption{CV+CS+FR}
    \label{fig:greenhouse_a_cvcsfr}
  \end{minipage}

  \caption{Result of the adaptation on Greenhouse A}
  \label{fig:greenhouse1_result}
\end{figure*}

Fig. \ref{fig:greenhouse1_result} is
the result of the adaptation on Greenhouse A.
In Fig. \ref{fig:greenhouse1_result},
while the results of the single-source training are
relatively noisy, the multi-source ones provide smoother outputs.
In particular, the plant regions are classified more accurately.
Qualitatively, we suppose that this level of accuracy is sufficient
for the operation of mobile robots.
For Greenhouse B and C, the results are shown in
Fig. \ref{fig:greenhouse2_result} and
\ref{fig:cucumber_result} in Appendix \ref{sec:appendix_qualitative_b_and_c}.
From Figs. \ref{fig:greenhouse2_result} and \ref{fig:cucumber_result},
we can see that accurate segmentation can be achieved in other environments,
and thus the knowledge that the proposed method transfers from the source datasets
is applicable in a variety of greenhouse environments.

\hypertarget{result-comparison}{%
  \subsection{Comparison to existing methods}
  \label{sec:result-result-comparison}}

\subsubsection{Baseline methods}

To evaluate the relative performance of our proposed method,
we conducted training with existing methods of
supervised learning as well as single-source and multi-source UDA for semantic segmentation.
Implementation details of the baseline methods
are described in Appendix \ref{sec:appendix_impl_details}.

\noindent
\textbf{Supervised learning methods}
We introduce a supervised support vector machine (SVM) on superpixel features
and a supervised DNN as supervised baselines.
The SVM method is equivalent to some existing segmentation methods
such as Sharifi et al. \cite{Sharifi2015}
and Lulio et al. \cite{Lulio2009}, where
an image is partitioned into small segments 
and then classified based on their hand-crafted features.
As a DNN model, we use the architecture that is the same as the proposed method.
We refer to the SVM method as \textit{SP-SVM}
and the DNN method as \textit{SP-DNN}. 
Implementation details of SP-SVM are in Sec. \ref{sec:appendix_feature_design}.


For evaluation of the supervised baselines, we conducted 10-fold cross-validation
with the 100 labeled images of Greenhouse A,
which are used as the test data in the experiments of the UDA methods.
The reported results are averages of
IoUs on the test fold of each of the 10 training trials.
Note that the training setting and the evaluation condition
are different from the UDA training.
This is due to the limited amount of manually labeled greenhouse images available.
Nevertheless, we report the results of the supervised baselines to demonstrate
baseline performances that we can expect from supervised learning.

\noindent
\textbf{UDA}
As existing UDA methods for semantic segmentation,
we introduce representative UDA methods in different generations,
i.e., confidence regularized self-training (CRST) \cite{Zou2019}
proposed in 2019,
Seg-Uncertainty \cite{Zheng2021} and ProDA \cite{Zhang2021} proposed in 2021.
They employ self-supervised learning with pseudo-labels,
which is a popular approach in current UDA methods for semantic segmentation.
Seg-Uncertainty and ProDA also involve
GAN-based training based on \cite{Tsai2018}
to further enhance the adaptation.
Seg-Uncertainty and ProDA are the original works
of the uncertainty-based loss weighting and the pseudo-label denoising
in our method, respectively.

We used the source code distributed by the authors
with modifications to apply them to our experiments, such as
adding the new datasets and converting the source labels to
the target label set defined in Table \ref{table:label_mapping}.
In terms of ProDA, we report the result of stage 1
proposed in \cite{Zhang2021}.
Although the original method employs knowledge distillation after stage 1
to boost the performance,
they reported the state-of-the-art performance at the time
with only the training of stage 1.
We, therefore, suppose the result of stage 1 on our setting
is sufficient as a baseline.
The rest of the training process follows the algorithm of each method.
Implementation details are in Appendix \ref{sec:appendix_pre-training}.

Although there are also multi-source UDA methods such as
Multi-source Domain Adaptation for Semantic Segmentation (MADAN) \cite{Zhao2019a},
we could not train it due to limitations of computational resources.
In Appendix \ref{sec:GAN},
we show results of image style transfer between
the source dataset and the target greenhouse dataset,
which is the first step of MADAN.


\subsubsection{Source datasets}

In the comparative studies of single-source UDA methods,
we report the results on Cityscapes (CS)
since it provided best results than CV and FR)
in all of the single-source baseline methods.
We suppose this may be because CS has a sufficient amount of images (about 3000)
for training models with a large capacity such as DeepLab v2,
while the other source datasets (CV and FR) have only a few hundreds.
On CV and FR, the model may have overfit to the training sets
due to the scarcity of the data, and thus resulted in poorer adaptation performances.
As there was no significant difference of adaptation performance
between the source datasets in the single-source setting
shown in Sec. \ref{results} (see Table \ref{table:experiment1}),
we believe it is reasonable to use CS to
demonstrate the performance of the baselines
on behalf of the three source datasets.

%
%

\subsubsection{Results}

Table \ref{table:experiment_comparative} shows the results.
\begin{table*}[tb]
  \centering
  \caption{Results of the baseline methods. CS is used in the single-source UDA baselines.
  \textbf{Bold} denotes the best performance among the UDA methods}
  \label{table:experiment_comparative}
  \begin{tabular}{ c c c c c }
    \toprule
    \multirow{2}[0]{*}{\textbf{Method}}
                                     & \multicolumn{3}{c}{\textbf{Class IoU}} & \multirow{2}[0]{*}{\textbf{mIoU}}                                    \\
                                     & \textbf{Plant}                         & \textbf{Artificial object}        & \textbf{Ground} &                \\
    \midrule
    SP-SVM*                          & 74.99                                  & 54.66                             & 53.54           & 61.06          \\
    SP-DNN*                          & 86.27                                  & 78.64                             & 88.49           & 81.68     \\
    \midrule 
    CRST \cite{Zou2019}              & 78.89                                  & 76.85                 & 65.10           & 73.62          \\
    Seg-Uncertainty \cite{Zheng2021} & 52.17                                  & 70.68                             & 27.29           & 50.04          \\
    ProDA \cite{Zhang2021}           & 56.15                                  & 60.81                             & 28.53           & 48.50          \\
    CV+CS+FR (Proposed)              & \textbf{80.71}                         & \textbf{78.21}                    & \textbf{72.68}  & \textbf{77.20} \\
    \bottomrule
  \end{tabular}
  \flushleft{\footnotesize{*For SP-SVM and SP-DNN, we report averages of 10-fold cross-validation using 100 labeled images.}}

\end{table*}
Overall, the proposed method outperformed the baseline methods
except for SP-DNN,
and thus exhibited the ability to
transfer knowledge from the datasets dissimilar to the target dataset.
Although the performance of the baseline methods might be improved with
extensive hyper-parameter search,
the results indicate that relying on a single source
is not effective enough even when using
state-of-the-art UDA methods in our task
where the source and target datasets
do not share their structure of the scenes.
Qualitative results are shown in Appendix \ref{sec:qualitative_baselines}.
While SP-DNN resulted in the best IoUs,
it also showed a clear tendency of overfitting due to the scarcity of training data.
Although increasing the training data will resolve the problem, 
it requires laborious manual annotation.
On the other hand, the proposed method enables training with a large amount of 
unlabeled images and provides a comparative performance.



Among the baseline UDA methods,
CRST resulted in better performance than others,
although they are reported to be better than CRST
\cite{Zheng2021,Zhang2021}.
We conjecture that adversarial learning negatively contributed to
these results.
ProDA and Seg-Uncertainty involve
adversarial learning in the training processes
to make the spatial structure in the output space
as close as possible between the source and the target datasets \cite{Tsai2018}.
Adversarial learning, including GANs \cite{Goodfellow2014}, 
is effective when the difference between the source and the target
stems mainly from the styles such as texture and color
and the datasets share similar structure.
For instance, GANs can be used to transfer the image style of source images
closer to the target images without significantly changing the image contents,
i.e., semantic class of corresponding image regions,
so that a segmenter for the target images can be trained
on the transferred source images in the target's style
with the ground truth labels.
However, when the structure of the scenes is different,
adversarial learning does not necessarily convey meaningful information.
In fact, as shown in Appendix \ref{sec:GAN},
the image contents are not preserved between
the original and the translated images
after CycleGAN-based image style transfer between CS/FR and Greenhouse A.
For example, some ground regions in CS are transferred to
green plant-like stuff,
and some plant regions are transferred to regions with sky-like
appearance (See Fig. \ref{fig:cyclegan_examples}).
In \cite{Zhao2019a},
the authors also pointed out that the structural differences of objects between
the source and the target datasets
led the training to poor performances on those object classes
(e.g., the source in target images are much taller than those in target images).
Those facts imply 
adversarial learning is not as effective
for adaptation under such structural differences
between the source and the target scenes.


\hypertarget{result-multi-source-merging-strategy}{%
  \subsection{Comparison of strategies to merge multi-source information}
  \label{sec:result-multi-source-merging-strategy}}

\begin{table*}[tb]
  \centering
  \caption{Comparison of multi-source merging strategies}
  \label{table:comparison_pseudo-label_merging}
  \begin{tabular}{cc}
    \toprule
    Majority & Unanimity (proposed) \\ \midrule
    70.03    & \textbf{77.20}       \\
    \bottomrule
  \end{tabular}
\end{table*}

To see the validity of the proposed pseudo-label generation
based on unanimity of the source models,
we conducted training with a different strategy,
namely majority-based pseudo-label generation.
In the majority-based strategy, a label is assigned to a pixel
when more than two source models of the three predict the pixel
as the same object class.

The results are shown in Table \ref{table:comparison_pseudo-label_merging}.
The training with the majority-based pseudo-labels resulted in
a significant drop in performance compared to the unanimity-based method.
We suppose that since the majority-based method
more aggressively involves pixels in the pseudo-labels,
more noise was included and it affected the training.
We can conclude that the pseudo-labels should be chosen carefully
through the strict unanimity criteria for better training.
Although the pseudo-labels generated in such a way become coarse,
the model trained with them provides reasonable predictions
as shown in the aforementioned experiments.
The validity of training with coarse labels is also reported in \cite{Zlateski2018}.

\hypertarget{result-pseudo-label-update}{%
  \subsection{Effect of updating the pseudo-labels}
  \label{sec:result-pseudo-label-update-strategy}}

We evaluate the effect of 
pseudo-label update strategies.
We compare the strategies as follows.

\begin{itemize}
\item \textbf{Initial} the initial pseudo-labels are used throughout the training.

\item \textbf{Update once} the pseudo-labels are updated after three rounds (i.e., 15 epochs).


\item \textbf{Update every round}
the pseudo-labels are updated every round after the first three rounds.
\end{itemize}

The results are shown in Table \ref{table:comparison_pseudo-label_update}.
\begin{table*}[b]
  \centering
  \caption{Results of the training with and without pseudo-label update}
  \label{table:comparison_pseudo-label_update}
  \begin{tabular}{ccc}
    \toprule
    Initial & Update once    & Update every round \\ \midrule
    75.44   & \textbf{77.20} & 74.82              \\
    \bottomrule
  \end{tabular}
\end{table*}
\textit{Update once} outperformed the other strategies.
We suppose the reason is as follows.
The initial pseudo-labels are sparse 
and many pixels are ignored in training.
Although it plays a crucial role to transfer primary knowledge
from multiple source datasets, 
using such labels throughout the training leads to a suboptimal result.
Updating the pseudo-labels with \textit{Update once} strategy
enables to exploit the knowledge learned through
the training with the initial pseudo-labels and involve more pixels in the training
to further improve the performance.

While the label update improved the performance,
updating the labels every round resulted in the worst IoU.
This is an expected result because the pseudo-labels
inevitably include wrong ones due to misclassifications
and the errors are accumulated as the pseudo-labels are updated
by the model trained with them.
This problem is also pointed out in \cite{Zhang2021}.
We, therefore, conclude that the pseudo-labels should be updated
once in the early stage of the training.

\hypertarget{result-ablation}{%
  \subsection{Ablation on pseudo-label noise suppression strategies}
  \label{sec:result-pseudo-label-noise-suppression-strategies}}

We conducted an ablation study
on the methods for noise suppression. 
We examined the methods as follows:
confidence thresholding, loss weighting, and prototype-based denoising.
The confidence thresholding is simply removing outputs
with a confidence value below a predefined threshold
from the pseudo-labels.
Here, the confidence threshold is set to 0.9 in all the settings.
The loss weighting and the prototype-based denoising are the methods
described in Sec. \ref{sec:model_training_loss_weighting} and
\ref{sec:model_training_prototype_denoising}, respectively.

We report the training results on Greenhouse A.
The results are shown in Table \ref{table:ablation_pseudo-label_denoising}.
\begin{table*}[tb]
  \centering
  \caption{Ablation on the pseudo-label noise suppression strategies}
  \label{table:ablation_pseudo-label_denoising}
  \begin{tabular}{cccc}
    \toprule
    Confidence threshold & Loss weighting \cite{Zheng2021} & Prototype-based denoising \cite{Zhang2021} & mIoU           \\ \midrule
                         &                                 &                                            & 73.95          \\
    \checkmark           &                                 &                                            & 74.66          \\
    \checkmark           & \checkmark                      &                                            & 75.89          \\
    \checkmark           &                                 & \checkmark                                 & 75.53          \\
    \checkmark           & \checkmark                      & \checkmark                                 & \textbf{77.20} \\
    \bottomrule
  \end{tabular}
\end{table*}
The strategies for suppressing the effect of noise in pseudo-labels
effectively improved the training performance.
The combination of the loss weighting and the prototype-based pseudo-label denoising
on top of the simple confidence thresholding resulted in the best performance.
We, therefore, adopted this setting to all the other training we report in this paper.

\hypertarget{result-alpha-sensitivity}{%
  \subsection{Parameter sensitivity analysis}
  \label{sec:result-parameter-sensitivity-analysis}}

\begin{table*}[tb]
  \centering
  \caption{Results of the training with different values of threshold $\alpha$}
  \label{table:comparison_confidence_threshold}
  \begin{tabular}{ccccc}
    \toprule
    $\alpha$ & 0.7   & 0.8   & 0.9            & 0.95  \\ \midrule
    mIoU     & 76.87 & 77.16 & \textbf{77.20} & 76.63 \\
    \bottomrule
  \end{tabular}
\end{table*}

We analyze the effect of the confidence threshold
in the pseudo-label generation.
We trained models with different thresholds on Greenhouse A dataset.
The results are shown in Table \ref{table:comparison_confidence_threshold}.
$\alpha=0.9$ resulted in the best performance.
However, the difference among the parameters
is less than one point and thus the method did not show
significant sensitivity to the parameter choice.

\subsection{Discussions}

Here, we list the limitations of the proposed method.

\begin{enumerate}
  \item \textit{The limitation of the pseudo-label generation strategy}

        In the proposed method, pseudo-labels are selected for training only
        when all the source models unanimously make predictions on the corresponding pixels.
        Although such a conservative strategy contributed to
        removing noise as shown in Sec. \ref{sec:result-multi-source-merging-strategy},
        it may also result in lack of valid labels
        if one of the source datasets do not agree with others. 
        This fact also hinders increasing the number of source datasets
        since the more the source models are, the harder
        it becomes to get agreement of all the source models
        and thus the produced pseudo-labels become too sparse.

  \item \textit{The limitation of applicable environments}

        In this paper, we confirmed that the proposed method is applicable to
        multiple greenhouses with a similar structure.
        We are now looking to apply our method to unstructured outdoor scenes.
        We have not confirmed that the same method also works for such scenes.
        We should, therefore, investigate the applicability of
        the proposed method to other scenes, e.g.,
        unstructured outdoor scenes and forests.

  \item \textit{The limitation of capability to learn novel object classes}

        There may be cases where the target dataset has classes
        that do not appear in the source datasets, e.g., objects specific to the environment.
        The proposed method is not capable of learning such novel objects.

\end{enumerate}
\section{CONCLUSIONS}

In this paper, we tackled a task of unsupervised domain adaptation from
multiple outdoor datasets to greenhouse datasets, to train a model
for scene recognition of agricultural mobile robots.
We considered using multiple publicly available datasets of outdoor images as source datasets
since it is difficult to prepare synthetic datasets of greenhouses 
as done in conventional work.
To effectively exploit the knowledge from the source datasets whose appearance and 
structure is very different from the target environment, 
we proposed a pseudo-label generation method that takes outputs from each source model
and selects only the labels that are unanimously predicted by all the source models.
By the combination of our pseudo-label generation method and conventional methods, 
training of semantic segmentation on the greenhouse datasets
is enabled without manual labeling of images.
This contributes to reducing the burden of applying a visual scene recognition system of 
mobile robots in greenhouses.
In fact, we have already applied the model trained with the proposed method 
to the recognition of a mobile robot for autonomous navigation \cite{Matsuzaki2022}.

\section*{Disclosure statement}

The authors report there are no competing interests to declare.

\section*{Funding}

This work is supported by Knowledge Hub Aichi Priority
Research Project (third term). The work of Shigemichi Matsuzaki is also supported
in part by the Leading Graduate School Program, “Innovative
program for training brain-science-information-architects by
analysis of massive quantities of highly technical information
about the brain,” by the Ministry of Education, Culture,
Sports, Science and Technology, Japan.

\section*{Authors' biography}

Shigemichi Matsuzaki  received the associate degree in engineering from
National College of Technology, Kumamoto College in 2016,
and the bachelor's and the master's degrees in engineering
from Toyohashi University of Technology, Aichi, Japan
in 2018 and 2020, respectively.
He is currently pursuing the Ph.D. degree 
at Toyohashi University of Technology.
His research interests include deep learning in robot perception,
mobile robots in unstructured environments, and mobile service robots.

Jun Miura received the B.Eng. degree in mechanical engineering, and
the M.Eng. and Dr.Eng. degrees in information engineering from The University of Tokyo,
Tokyo, Japan, in 1984, 1986, and 1989, respectively. In 1989, he joined the Department of
Computer-Controlled Mechanical Systems, Osaka
University, Suita, Japan. Since April 2007, he has
been a Professor with the Department of Computer
Science and Engineering, Toyohashi University of
Technology, Toyohashi, Japan. From March 1994 to February 1995, he was a
Visiting Scientist with the Computer Science Department, Carnegie Mellon
University, Pittsburgh, PA. He has published over 240 articles in international
journal and conferences in the areas of intelligent robotics, mobile service
robots, robot vision, and artificial intelligence. He received several awards,
including the Best Paper Award from the Robotics Society of Japan, in 1997,
the Best Paper Award Finalist at ICRA-1995, and the Best Service Robotics
Paper Award Finalist at ICRA-2013

Hiroaki Masuzawa received the associate degree in engineering from
National College of Technology, Hachinohe College in 2006,
received the bachelor's and the master's degree in engineering from
Toyohashi University, Aichi, Japan in 2008 and 2010, respectively.
He is currently a project
research associate of Department of Computer Science and Engineering,
Toyohashi University of Technology.  His research interests include
application of artificial intelligence to agricultural robots.

\bibliographystyle{tADR}
\bibliography{Writing-journal2020}

\appendices

\clearpage
\section{Network architecture}
\label{sec:network_architecture}

\begin{figure*}[tb]
  \centering

  \begin{minipage}[b]{0.47\hsize}
    \includegraphics[width=\hsize]{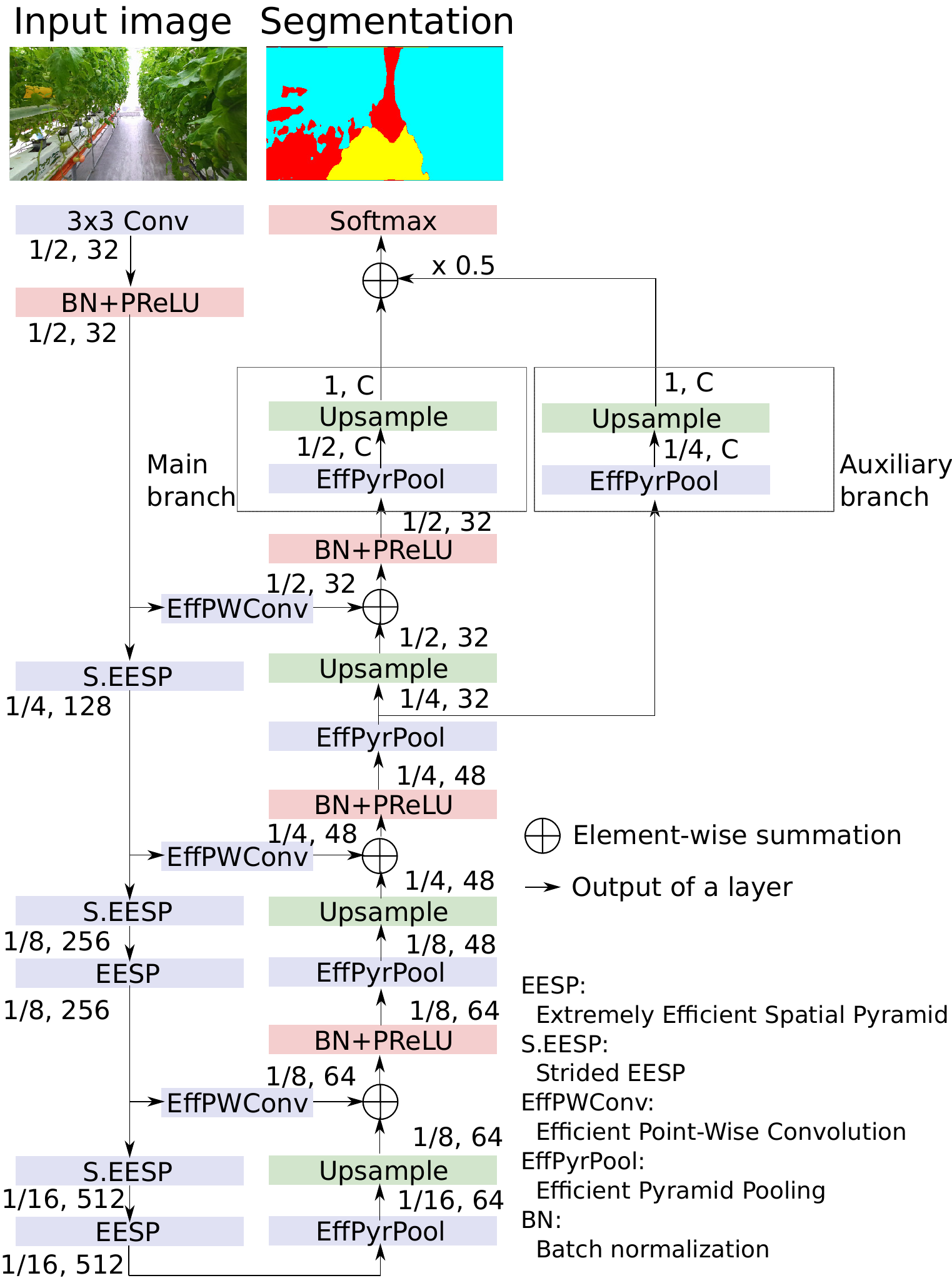}
    \subcaption{The entire network}
    \label{fig:network_detail_entire_network}
  \end{minipage}
  \begin{minipage}[b]{0.52\hsize}
    \includegraphics[width=\hsize]{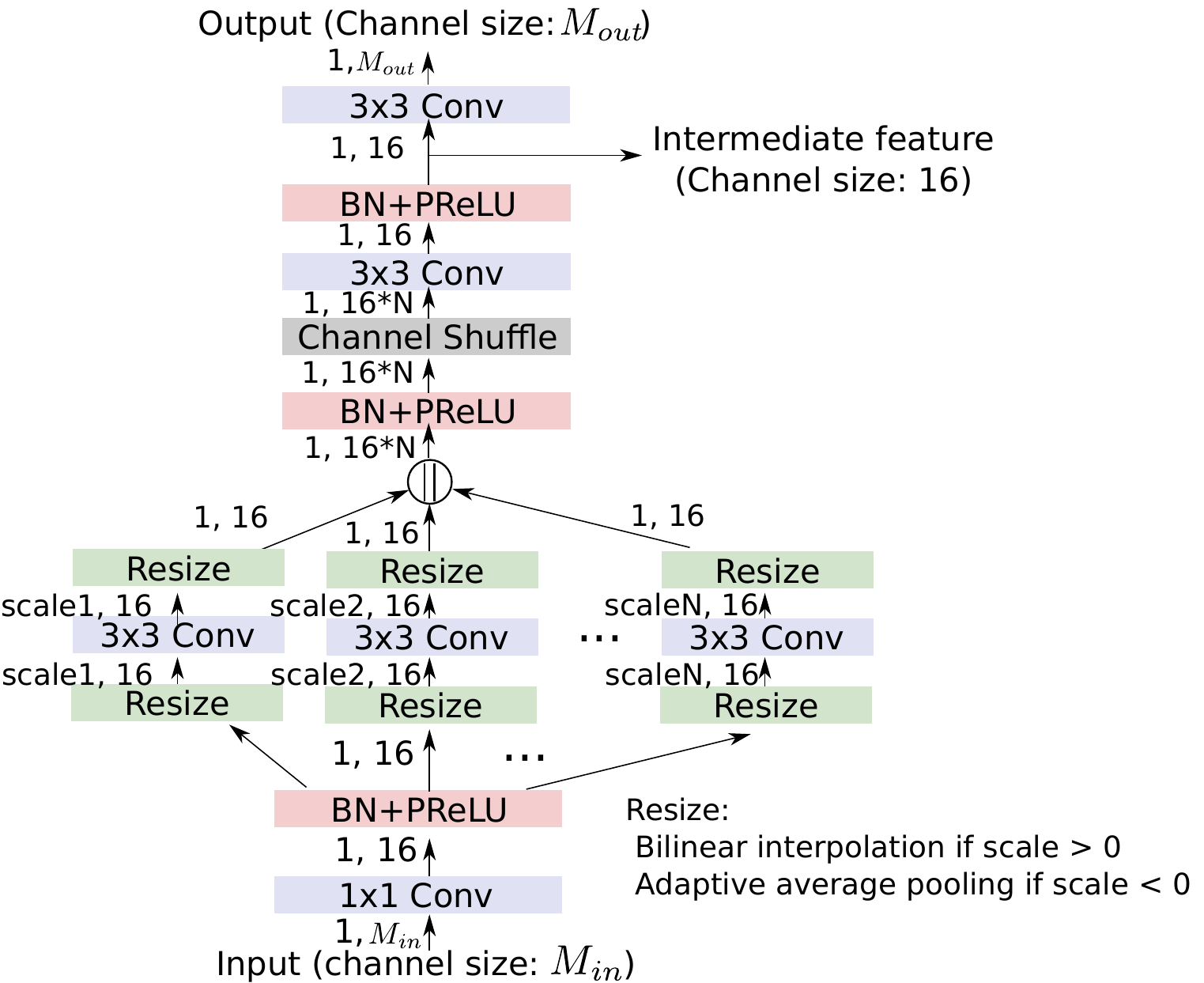}
    \subcaption{Efficient Pyramid Pooling (EffPyrPool)}
    \label{fig:network_detail_effpyrpool}
  \end{minipage}

  \caption{A detailed diagram of our architecture.
    The numbers beside each layer
    shows the spatial resolution relative to the input image and
    the channel size of the output of the layer, respectively.
    Based on ESPNetv2 \cite{Mehta2019}, an auxiliary segmentation branch
    is added to estimate pixel-wise uncertainty.
    For the detail of each network layer, refer to \cite{Mehta2019}.}
  \label{fig:network_detail}
\end{figure*}

A diagram of the network architecture is
shown in Fig. \ref{fig:network_detail_entire_network}.
Our network is based on ESPNetv2 \cite{Mehta2019},
a computationally efficient semantic segmentation network.
It consists of modules such as Extremely Efficient Spatial Pyramid (EESP),
Efficient Point-Wise Convolution (EffPWConv),
Efficient Pyramid Pooling (EffPyrPool).
For detailed descriptions of those modules,
we refer the readers to \cite{Mehta2019}.
An auxiliary segmentation branch
is attached to the middle of the main network of ESPNetv2
for estimating uncertainty used in the loss weighting \cite{Zheng2021}.

\section{Analysis of the prediction distributions}
\label{sec:appendix_pred_dist}

In Table \ref{table:prediction_distribution} , we show the distributions of the predictions
by each of the source models over the target classes conditioned on the source classes,
as well as the target label ID of the maximum probability for each source class and
the actual target label ID used in the label conversion
to see the validity of the heuristic label mappings.
The distributions are calculated using the labeled test images of Greenhouse A.
\begin{table*}[tb]
  \caption{Prediction distribution.
    1: Plant, 2: Artificial object, 3: Ground, 4: Other (not considered in the distribution analysis),
    in the target label space.}
  \label{table:prediction_distribution}
  \begin{minipage}[b]{\hsize}\centering
    \footnotesize
    \subcaption{CamVid}
    \label{table:prediction_distribution_camvid}
    \vspace{-5pt}
    \begin{tabular}{ccccccccccccc}
      \toprule
              & \myrot{Sky} & \myrot{Building} & \myrot{Pole} & \myrot{Road} & \myrot{Pavement} & \myrot{Tree} & \myrot{Sign symbol} & \myrot{Fence} & \myrot{Car} & \myrot{Pedestrian} & \myrot{Bicyclist} & \myrot{Road marking} \\
      \midrule
      1       & 1.72        & 19.58            & 11.87        & 0.64         & 0.98             & 95.30        & 26.39               & 93.07         & 0.00        & 0.00               & 36.00             & 28.63                \\
      2       & 92.01       & 80.00            & 71.27        & 38.33        & 37.80            & 4.70         & 73.61               & 6.93          & 80.98       & 98.25              & 61.33             & 59.29                \\
      3       & 6.27        & 0.42             & 16.87        & 61.03        & 61.23            & 0.00         & 0.00                & 0.00          & 19.02       & 1.75               & 2.67              & 12.09                \\
      \midrule
      max     & 2           & 2                & 2            & 3            & 3                & 1            & 2                   & 1             & 2           & 2                  & 2                 & 2                    \\
      $\xi_i$ & 4           & 2                & 2            & 3            & 3                & 1            & 2                   & 2             & 2           & 4                  & 4                 & 2                    \\
      \bottomrule
    \end{tabular}
  \end{minipage} \\

  \vspace{10pt}

  \begin{minipage}[b]{\hsize}\centering
    \footnotesize
    \subcaption{Cityscapes}
    \label{table:prediction_distribution_cityscapes}
    \vspace{-5pt}
    \begin{tabular}{cp{0.025\textwidth}p{0.025\textwidth}p{0.025\textwidth}p{0.025\textwidth}p{0.025\textwidth}p{0.025\textwidth}p{0.025\textwidth}p{0.025\textwidth}p{0.025\textwidth}p{0.025\textwidth}p{0.025\textwidth}p{0.025\textwidth}p{0.025\textwidth}p{0.025\textwidth}p{0.025\textwidth}p{0.025\textwidth}p{0.025\textwidth}p{0.025\textwidth}p{0.025\textwidth}}
      \toprule
              & \myrot{Road} & \myrot{Sidewalk} & \myrot{Building} & \myrot{Wall} & \myrot{Fence} & \myrot{Pole} & \myrot{Traffic light} & \myrot{Traffic sign} & \myrot{Vegetation} & \myrot{Terrain} & \myrot{Sky} & \myrot{Person} & \myrot{Rider} & \myrot{Car} & \myrot{Truck} & \myrot{Bus} & \myrot{Train} & \myrot{Motorcycle} & \myrot{Bicycle} \\
      \midrule
      1       & 0.00         & 6.19             & 0.54             & 0.00         & 4.55          & 1.18         & 0.00                  & 0.00                 & 83.02              & 0.00            & 0.00        & 0.00           & 0.00          & 0.00        & 0.00          & 0.00        & 0.00          & 0.00               & 28.81           \\
      2       & 54.28        & 21.12            & 99.46            & 0.00         & 90.00         & 81.25        & 0.00                  & 0.00                 & 16.76              & 0.00            & 0.00        & 100.0          & 0.00          & 85.64       & 0.00          & 100.0       & 0.00          & 0.00               & 69.49           \\
      3       & 45.72        & 72.69            & 0.00             & 0.00         & 5.45          & 17.57        & 0.00                  & 0.00                 & 0.22               & 0.00            & 0.00        & 0.00           & 0.00          & 14.36       & 0.00          & 0.00        & 0.00          & 0.00               & 1.69            \\
      \midrule
      max     & 2            & 3                & 2                & -            & 2             & 2            & -                     & -                    & 1                  & -               & -           & 2              & -             & 2           & -             & 2           & -             & -                  & 2               \\
      $\xi_i$ & 3            & 3                & 2                & 2            & 2             & 2            & 2                     & 2                    & 1                  & 3               & 4           & 4              & 4             & 2           & 2             & 2           & 2             & 2                  & 2               \\
      \bottomrule
    \end{tabular}
  \end{minipage} \\

  \vspace{10pt}

  \begin{minipage}[b]{\hsize}\centering
    \small
    \subcaption{Prediction distribution (Freiburg Forest)}
    \label{table:prediction_distribution_forest}
    \vspace{-5pt}
    \begin{tabular}{cccccc}
      \toprule
              & \myrot{Road} & \myrot{Grass} & \myrot{Tree} & \myrot{Sky} & \myrot{Obstacle} \\
      \midrule
      1       & 1.15         & 48.73         & 43.63        & 0.09        & 0.00             \\
      2       & 18.83        & 28.54         & 50.06        & 96.63       & 0.00             \\
      3       & 80.02        & 22.72         & 6.30         & 3.28        & 0.00             \\
      \midrule
      max     & 3            & 1             & 2            & 2           & -                \\
      $\xi_i$ & 3            & 1             & 1            & 4           & 2                \\
      \bottomrule
    \end{tabular}
  \end{minipage}

\end{table*}
In the majority of the source classes, the target label of the maximum probability
is the same as the one assigned heuristically.
``Fence'' in CamVid, ``Road'' in Cityscapes, and
``Tree'' in Freiburg Forest were assigned a target label different from
the one with the maximum probability.
In the model trained with Cityscapes,
some source classes did not appear in the prediction on the target images.

Note that these distributions are calculated with the ground truth labels of the target images,
which we do not expect to have in real environments,
and thus this information is unable.
In practice, as we show in the experiments,
the definition of the mappings based on the heuristics
resulted in good performances.

\section{Policy of manual annotation on the test images}
\label{sec:label_policy}

For the evaluation of our method,
we created test datasets for Greenhouse A, B, and C
with manually annotated labels.
All annotation was done by the first author,
who did not have specific experience of pixel-level annotation on image datasets.

We adopted coarse annotation to reduce the time of manual annotation.
For example, small regions of objects other than plants that
can be seen through the plant rows 
are annotated as plants, rather than the object classes 
that the regions actually belong to.
The aim our method is to train the model for scene recognition in robot navigation
to identify the object class of the regions in an image,
especially regions of plants covering the paths, and to make a decision whether
to traverse the object based on the object class.
For this purpose, we suppose that pixel-level precision is not necessary.
The model should rather be able to recognize the presence of plant etc.
We, therefore, gave region-wise annotation rather than labels with pixel-level precision,
so that the ability of recognizing the object class of image regions can be evaluated.

\section{Qualitative evaluation on Greenhouse B and C}
\label{sec:appendix_qualitative_b_and_c}

\begin{figure*}[tb]
  \raggedleft
  \begin{minipage}{0.45\hsize}
    \centering\includegraphics[width=\hsize]{legend.pdf}
  \end{minipage}\\

  \centering
  \begin{minipage}{0.16\hsize}
    \centering\includegraphics[width=\hsize]{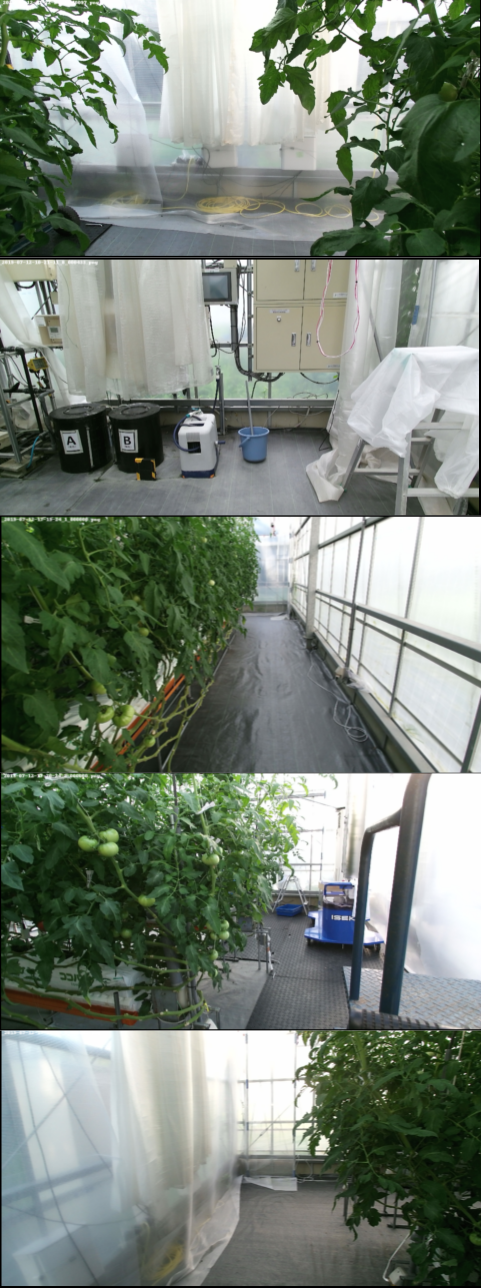}
    \subcaption{Camra image}
    \label{fig:greenhouse_b_camera}
  \end{minipage}
  \begin{minipage}{0.16\hsize}
    \centering\includegraphics[width=\hsize]{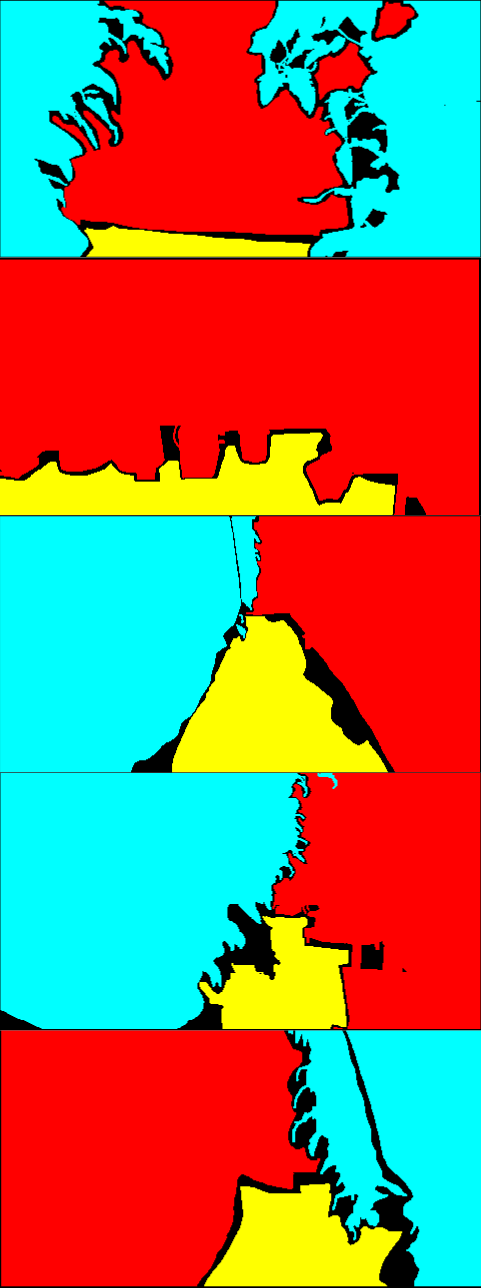}
    \subcaption{Ground truth}
    \label{fig:greenhouse_b_gt}
  \end{minipage}
  \begin{minipage}{0.16\hsize}
    \centering\includegraphics[width=\hsize]{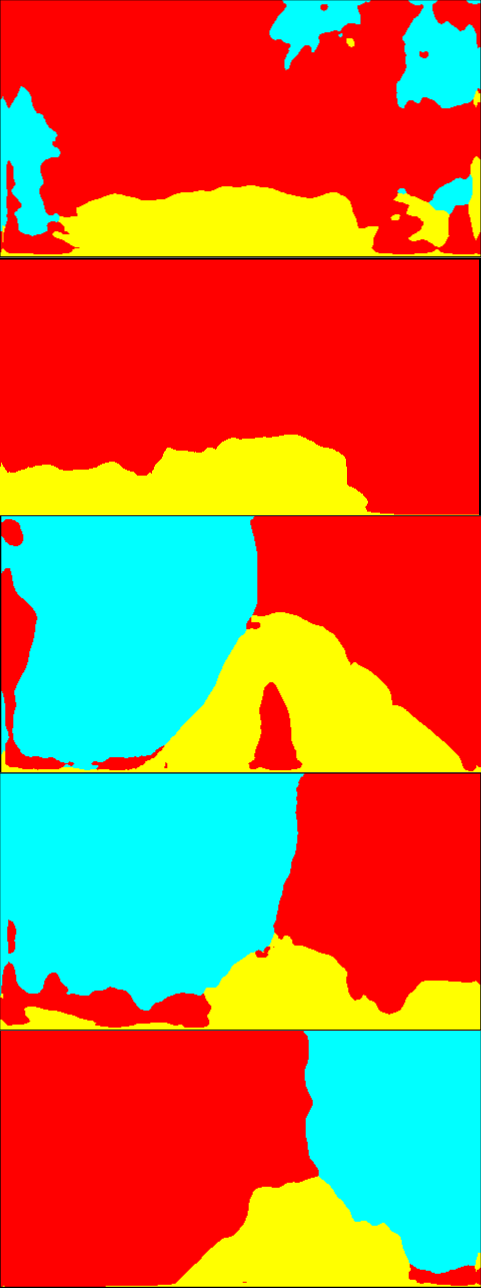}
    \subcaption{CV}
    \label{fig:greenhouse_b_cv}
  \end{minipage}
  \begin{minipage}{0.16\hsize}
    \centering\includegraphics[width=\hsize]{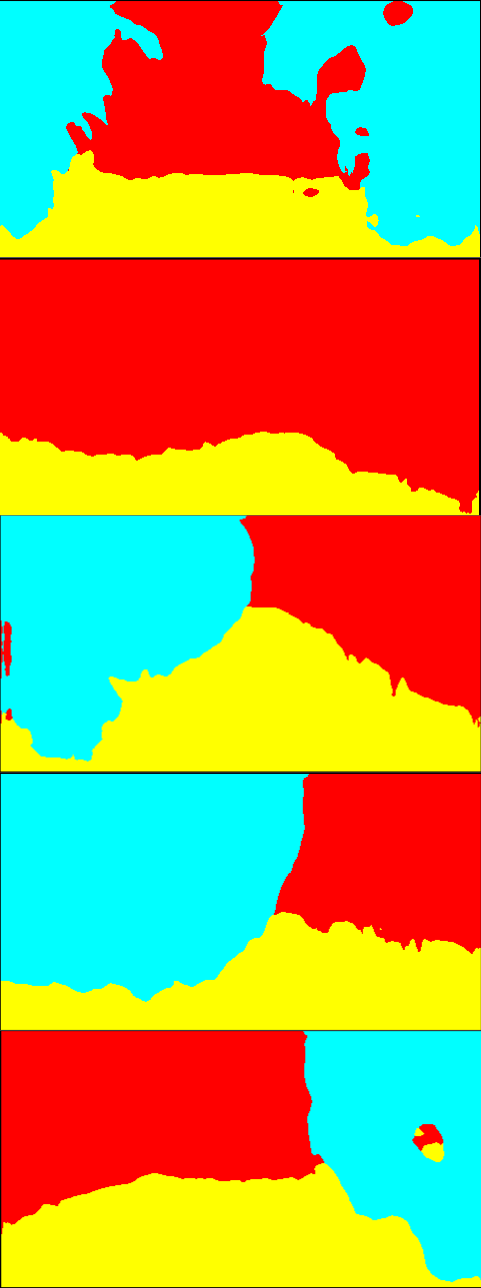}
    \subcaption{CS}
    \label{fig:greenhouse_b_cs}
  \end{minipage}
  \begin{minipage}{0.16\hsize}
    \centering\includegraphics[width=\hsize]{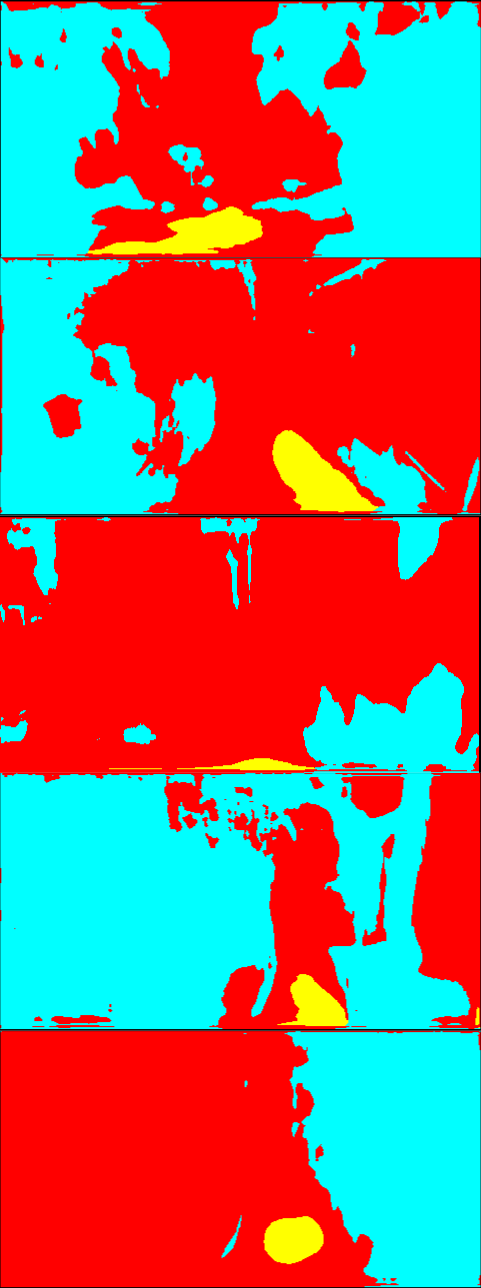}
    \subcaption{FR}
    \label{fig:greenhouse_b_fr}
  \end{minipage}
  \begin{minipage}{0.16\hsize}
    \centering\includegraphics[width=\hsize]{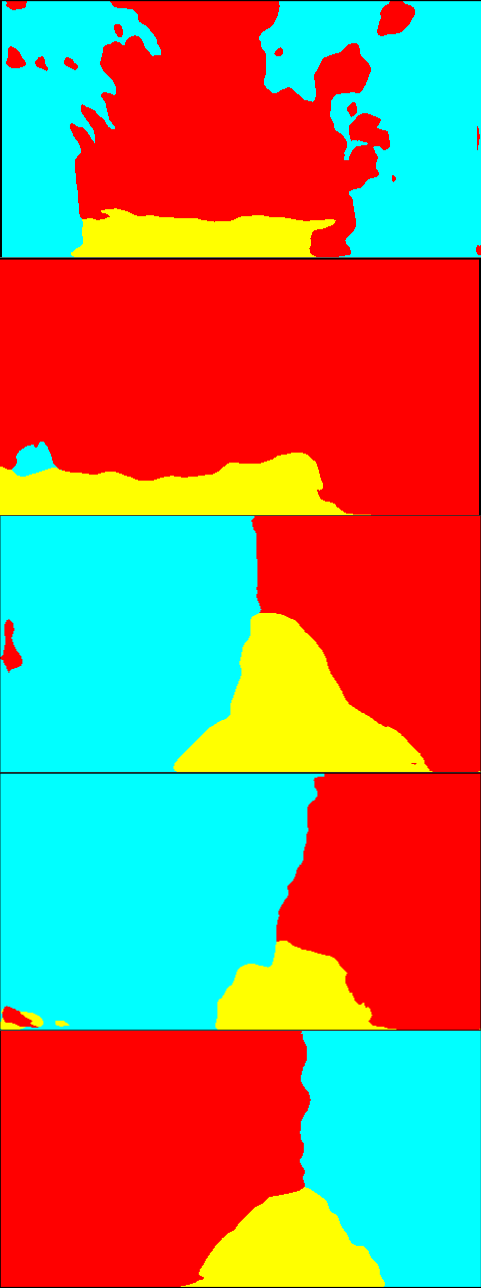}
    \subcaption{CV+CS+FR}
    \label{fig:greenhouse_b_cvcsfr}
  \end{minipage}

  \caption{Result of the adaptation on Greenhouse B}
  \label{fig:greenhouse2_result}

\end{figure*}

\begin{figure*}[tb]
  \raggedleft
  \begin{minipage}{0.45\hsize}
    \centering\includegraphics[width=\hsize]{legend.pdf}
  \end{minipage}\\

  \centering
  \begin{minipage}{0.16\hsize}
    \centering\includegraphics[width=\hsize]{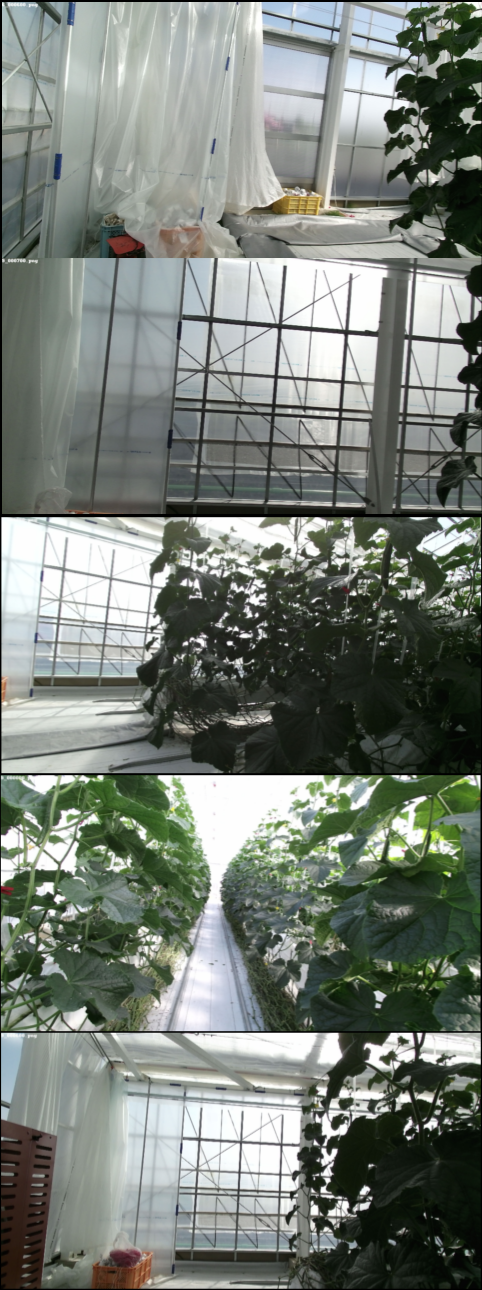}
    \subcaption{Camra image}
    \label{fig:greenhouse_c_camera}
  \end{minipage}
  \begin{minipage}{0.16\hsize}
    \centering\includegraphics[width=\hsize]{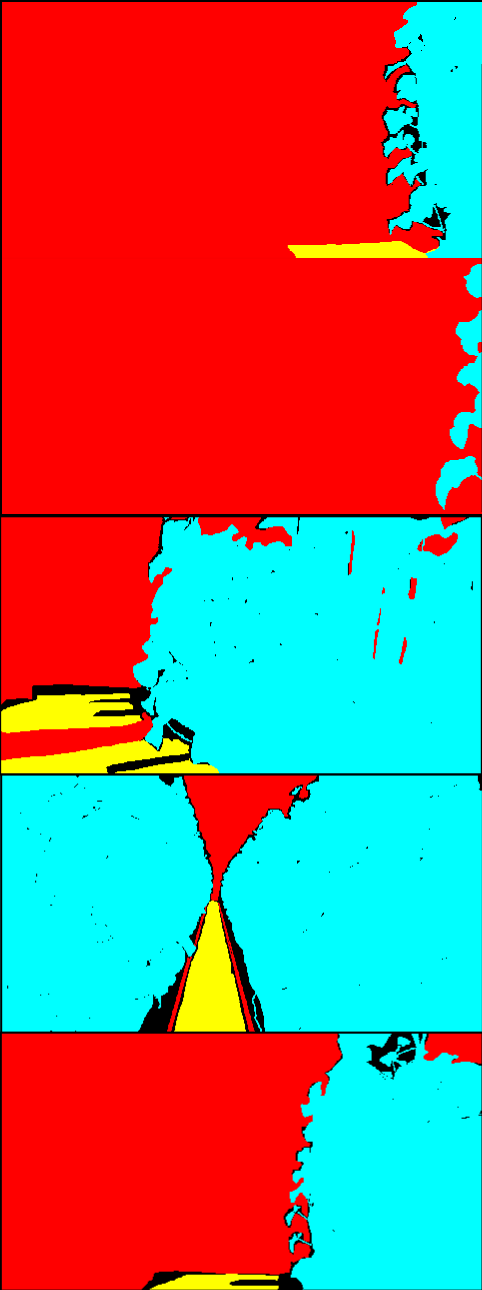}
    \subcaption{Ground truth}
    \label{fig:greenhouse_c_gt}
  \end{minipage}
  \begin{minipage}{0.16\hsize}
    \centering\includegraphics[width=\hsize]{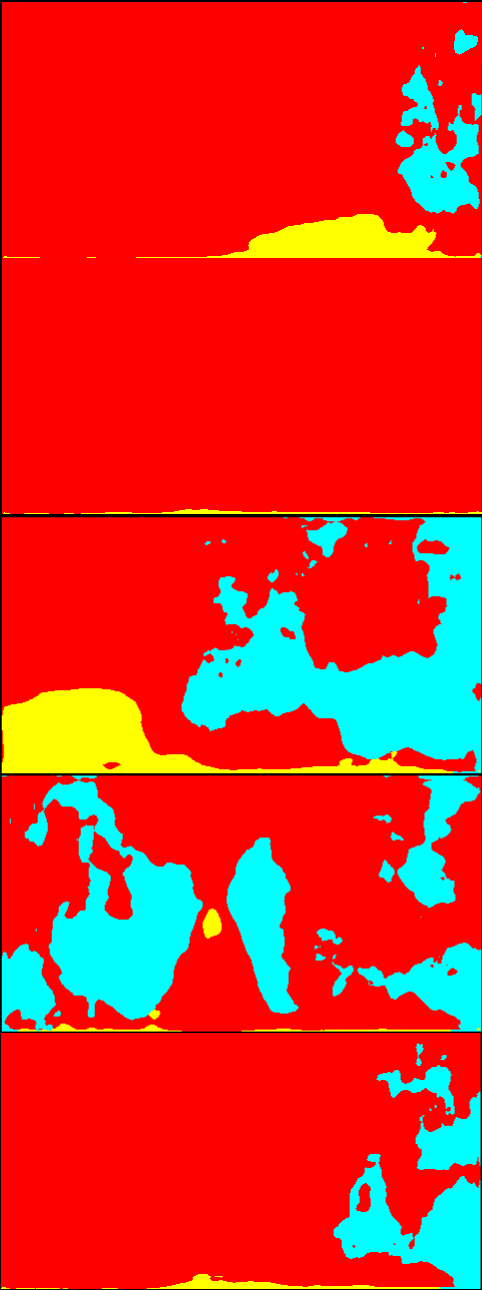}
    \subcaption{CV}
    \label{fig:greenhouse_c_cv}
  \end{minipage}
  \begin{minipage}{0.16\hsize}
    \centering\includegraphics[width=\hsize]{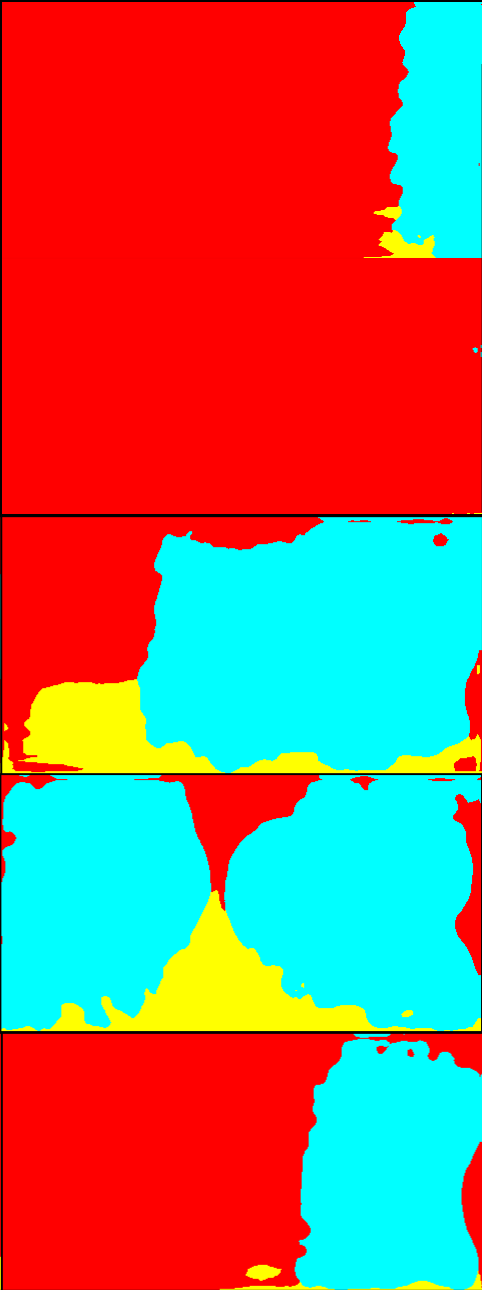}
    \subcaption{CS}
    \label{fig:greenhouse_c_cs}
  \end{minipage}
  \begin{minipage}{0.16\hsize}
    \centering\includegraphics[width=\hsize]{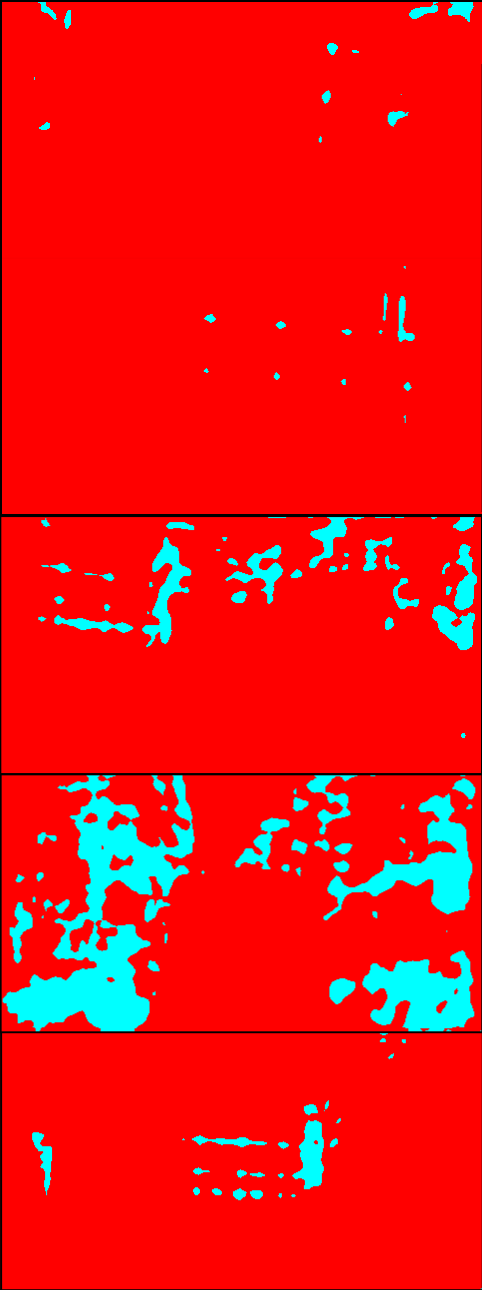}
    \subcaption{FR}
    \label{fig:greenhouse_c_fr}
  \end{minipage}
  \begin{minipage}{0.16\hsize}
    \centering\includegraphics[width=\hsize]{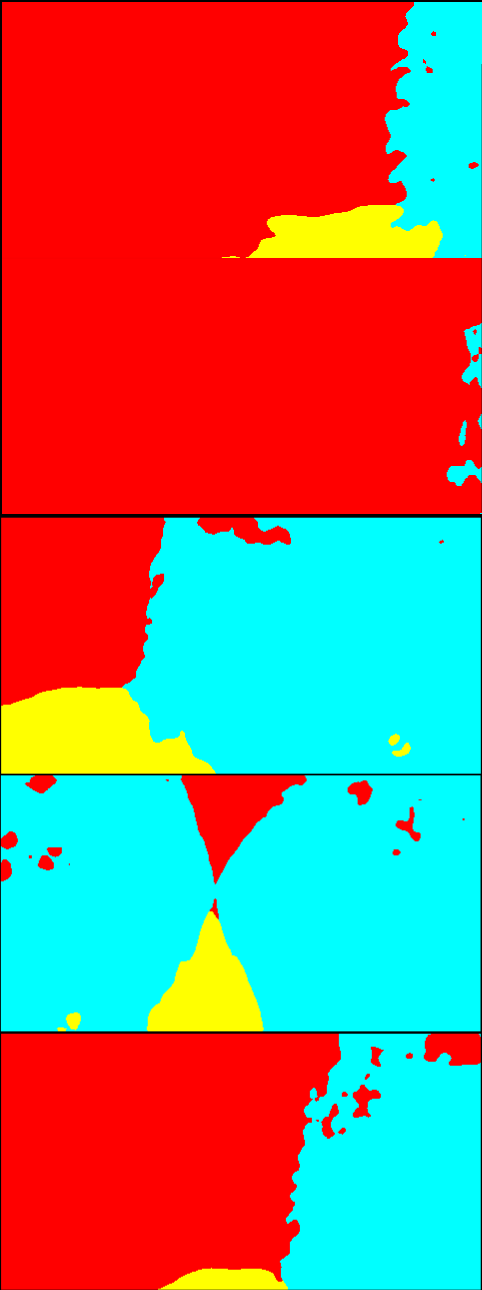}
    \subcaption{CV+CS+FR}
    \label{fig:greenhouse_c_cvcsfr}
  \end{minipage}

  \caption{Result of the adaptation on Greenhouse C}
  \label{fig:cucumber_result}

\end{figure*}

Fig. \ref{fig:greenhouse2_result} and \ref{fig:cucumber_result}
show the qualitative results of the adaptation to
Greenhouse B and C, respectively.
Similar to the results on Greenhouse A shown in \ref{results},
more smooth and noise-less segmentation is achieved by
the proposed method.

\section{Implementation details of the baseline methods}
\label{sec:appendix_impl_details}

\subsection{Description of the feature for SP-SVM}
\label{sec:appendix_feature_design}

We used the SVM model implemented in scikit-learn library \cite{Pedregosa2011}.
The Radial Basis Function (RBF) kernel was used with the default parameters
and one-versus-one decision function was chosen.
For superpixel generation, we used a method by Felzenszwalb et al. \cite{Felzenszwalb2004}
implemented in scikit-image library \cite{VanDerWalt2014}.

The feature design we adopt is based on the one used in a paper \cite{Kim2007}.
It consists of color features and texture features.
The color features are mean RGB and HSV values,
histograms of hue and saturation values.
The texture features are calculated via applying
filters in LM filter bank \cite{Leung2001}.
The filter responses are summarized by averaging and
taking maximum within the superpixel and concatenated.
In addition to the original feature design in \cite{Kim2007},
we added location features which are mean of $x$ and $y$ coordinates
to exploit the spatial information.

\begin{table*}[tb]
  \centering
  \caption{Description of the feature for SP-SVM}
  \label{table:feature_description}
  \begin{tabular}{  c  c  c  }
    \toprule
    Type       & Description                                                         & Dim. \\
    \midrule
    RGB value  & RGB mean                                                            & 3    \\
    HSV value  & RGB mean in HSV color space                                         & 3    \\
    Hue        & Histogram of hue                                                    & 8    \\
    Saturation & Histogram of saturation                                             & 5    \\
    LM average & Average filter responses from LM filter set                         & 18   \\
    LM Maximum & Histogram of maximum filter responses                               & 18   \\
    Location   & Mean x / y coordinate values normalized by the image width / height & 2    \\
    \midrule
               & Total number of dimension                                           & 57   \\
    \bottomrule
  \end{tabular}
\end{table*}

\subsection{Training details of the UDA baselines}
\label{sec:appendix_pre-training}

\subsubsection{Network architecture}

DeepLab v2 \cite{Chen2018-deeplab} with
ResNet101 backbone \cite{He2016} is used as the semantic segmentation model
in the original work of all the baseline methods of UDA.
Although it is different from our model, we also adopt DeepLab v2
in the baselines to keep their implementation as much as possible.
CRST and ProDA are trained on an NVIDIA Quadro RTX 8000,
and Seg-Uncertainty is trained on an NVIDIA GeForce 1080Ti with 11GB memory.

\subsubsection{CRST}

In CRST \cite{Zou2019},
the source models are pre-trained by supervised training
and we follow the implementation.
We first train a model with a source dataset
with its original label set for 200 epochs.
We then replace the final classification layer with $C$-class
classifier where $C$ denotes the number of object classes in the target dataset.
The model with the replaced classifier
is fine-tuned with the source labels converted to
the target label set defined in Table \ref{table:label_mapping}
for 50 epochs.
Initial learning rate is $1\times 10^{-4}$ for the final classification layer
and $1\times 10^{-3}$ for the rest of the layers.
The polynomial learning rate scheduling \cite{Mishra2019}
is used with the power of 0.9.
Other hyper-parameters are unchanged from their source code.

\subsubsection{Seg-Uncertainty}

We train Seg-Uncertainty \cite{Zheng2021} with MRNet \cite{Zheng2020a}.
MRNet first trains, as a warm-up, a segmentation network with both
the source and the target datasets via
adversarial training at the segmentation output
which was originally proposed in \cite{Tsai2018}.
Other hyper-parameters are unchanged from their source code.

\subsubsection{ProDA}

ProDA \cite{Zhang2021} follows the same warm-up strategy as Seg-Uncertainty.
We, therefore, use the same warm-up models trained in Seg-Uncertainty in ProDA.
Although ProDA and Seg-Uncertainty both use DeepLabv2 as a segmentation network,
the architecture of the segmentation networks is slightly different
between Seg-Uncertainty and ProDA, such as the number of channels of
the intermediate feature map.
We modified the source code of ProDA to adjust to the network.
Other hyper-parameters are unchanged from their source code.

%

\section{Results of the baseline methods}
\label{sec:qualitative_baselines}

\subsection{Qualitative evaluation of the UDA methods}
\label{sec:qualitative_uda}

\begin{figure*}[tb]
  \raggedleft
  \begin{minipage}{0.45\hsize}
    \centering\includegraphics[width=\hsize]{legend.pdf}
  \end{minipage}\\

  \centering
  \begin{minipage}{0.16\hsize}
    \centering\includegraphics[width=\hsize]{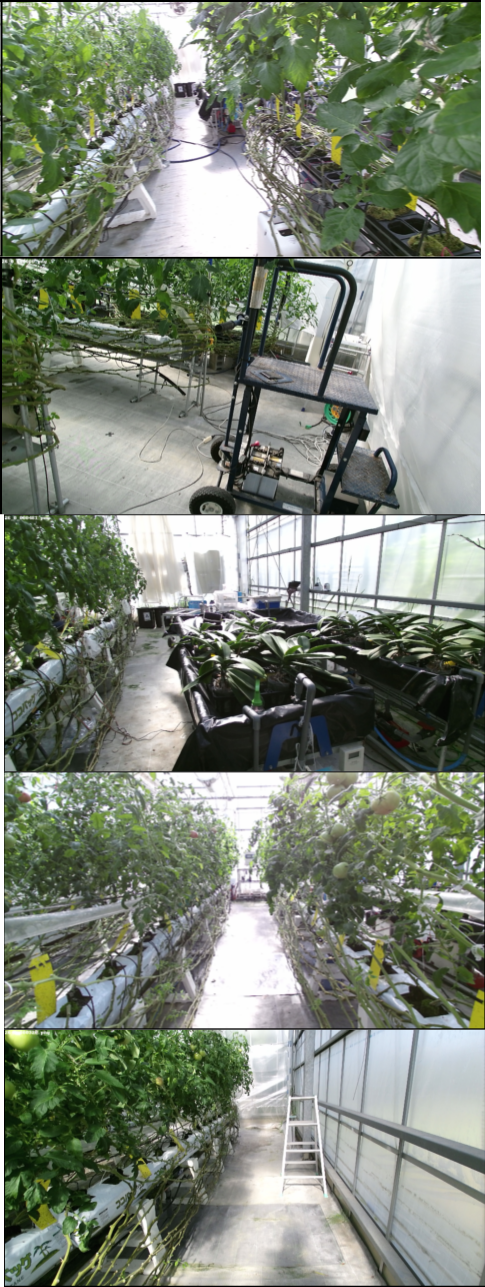}
    \subcaption{Camra image}
    \label{fig:greenhouse_a_camera_baseline}
  \end{minipage}
  \begin{minipage}{0.16\hsize}
    \centering\includegraphics[width=\hsize]{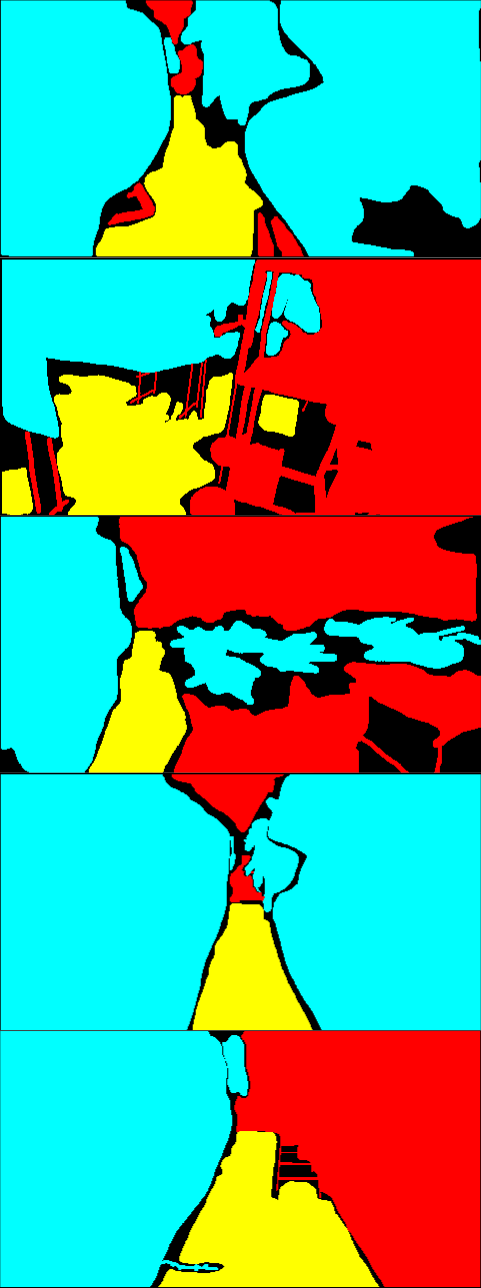}
    \subcaption{Ground truth}
    \label{fig:greenhouse_a_gt_baseline}
  \end{minipage}
  \begin{minipage}{0.1523\hsize}
    \centering\includegraphics[width=\hsize]{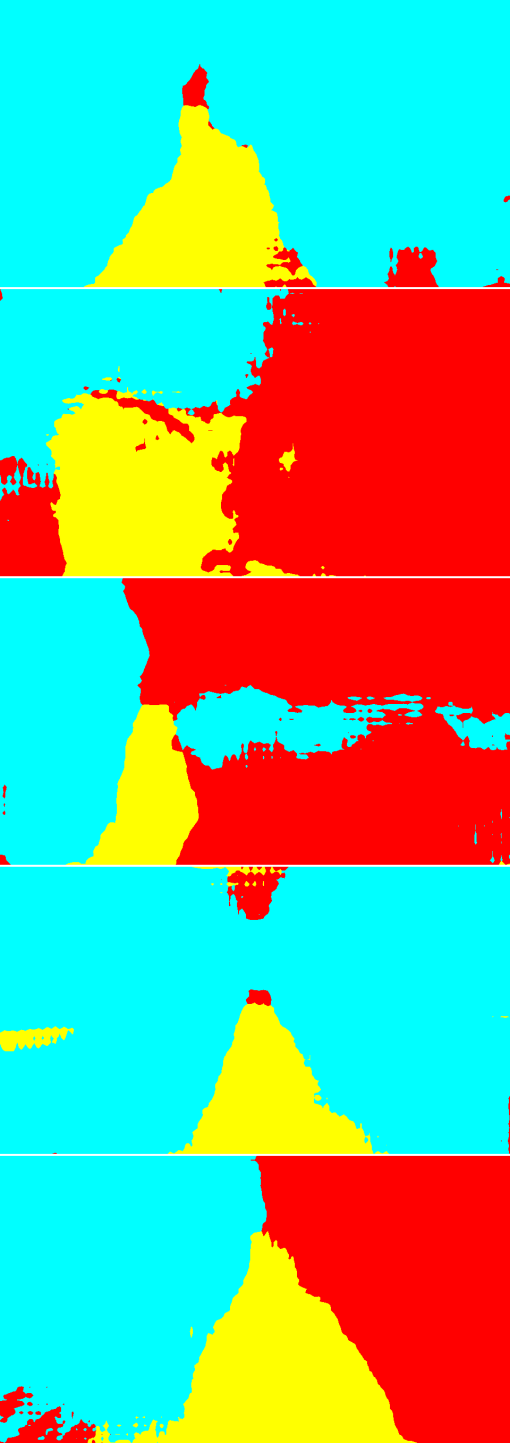}
    \subcaption{CRST \cite{Zou2019}}
    \label{fig:greenhouse_a_crst}
  \end{minipage}
  \begin{minipage}{0.1523\hsize}
    \centering\includegraphics[width=\hsize]{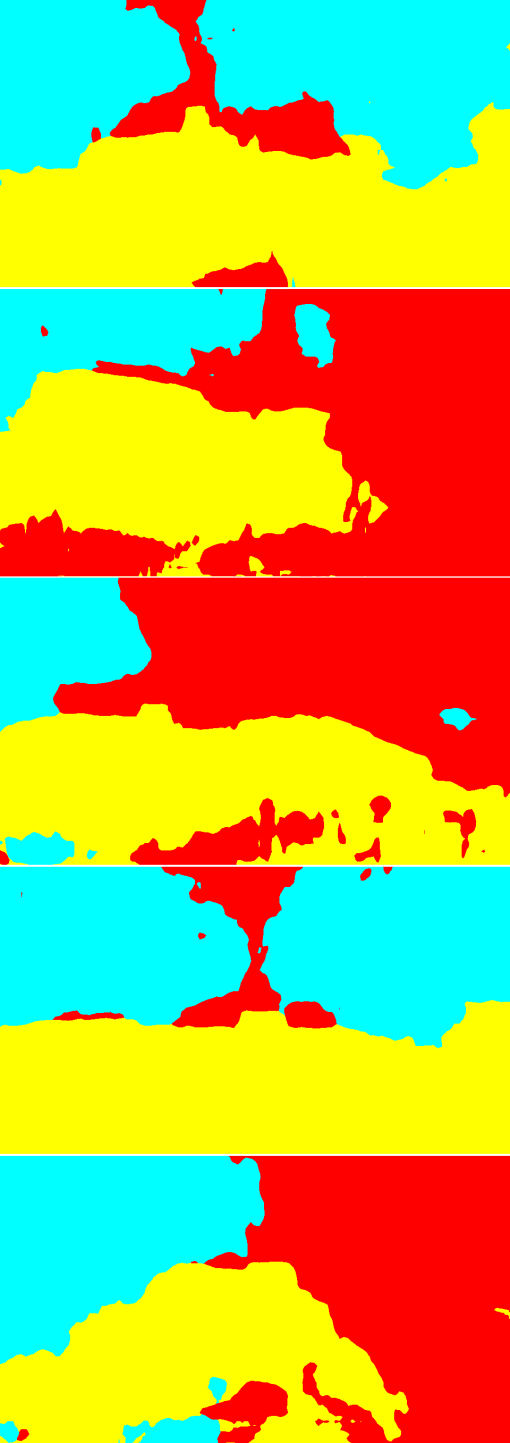}
    \subcaption{Seg-Uncert. \cite{Zheng2021}}
    \label{fig:greenhouse_a_su}
  \end{minipage}
  \begin{minipage}{0.1523\hsize}
    \centering\includegraphics[width=\hsize]{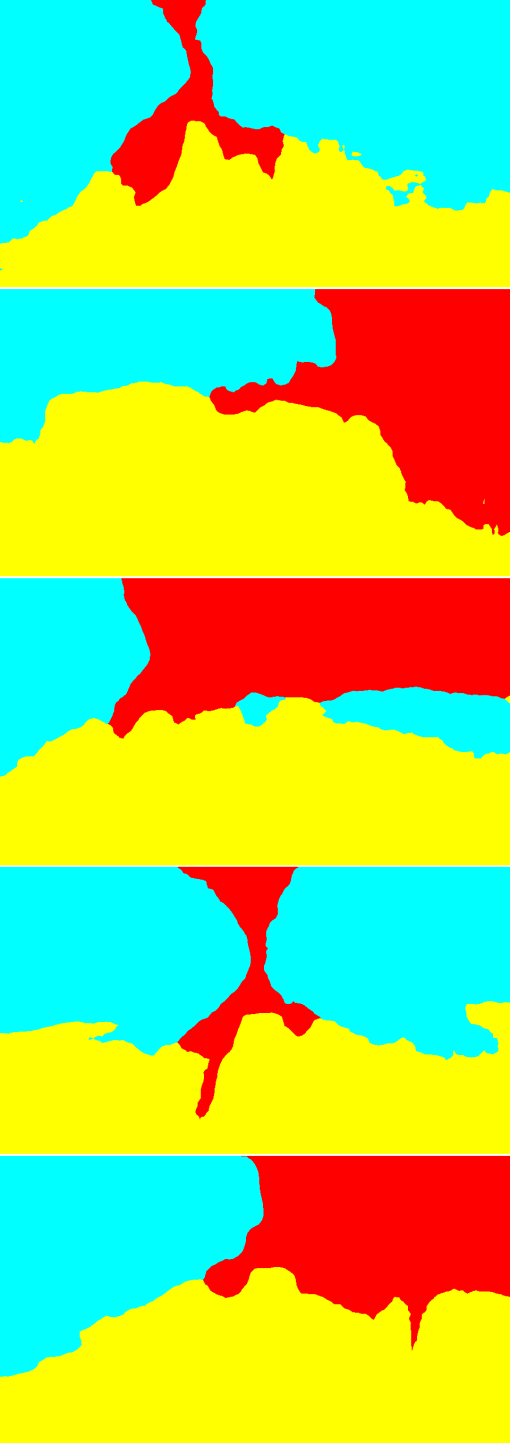}
    \subcaption{ProDA \cite{Zhang2021}}
    \label{fig:greenhouse_a_proda}
  \end{minipage}
  \begin{minipage}{0.16\hsize}
    \centering\includegraphics[width=\hsize]{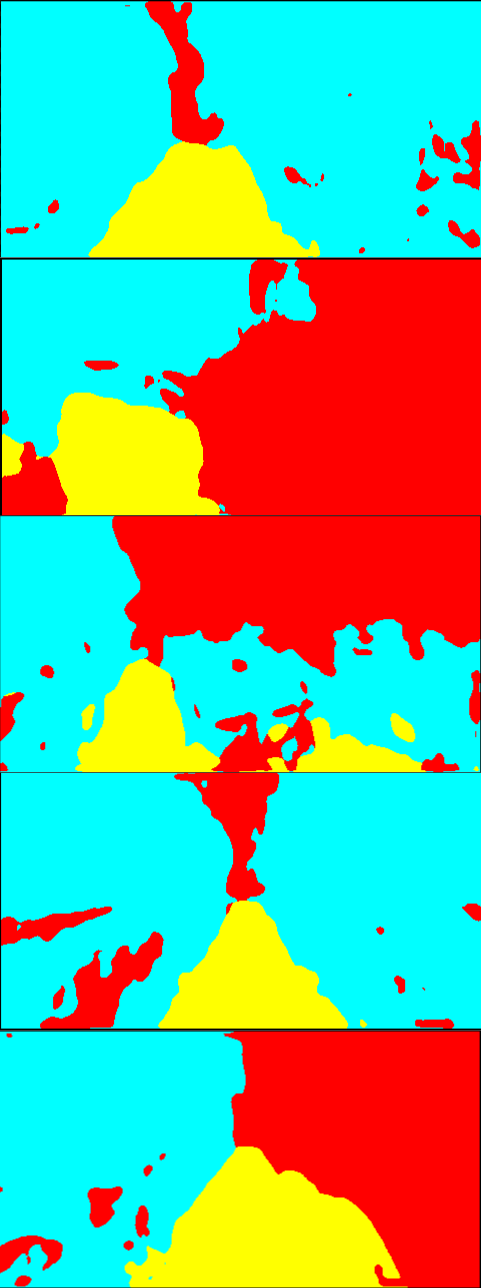}
    \subcaption{CV+CS+FR}
    \label{fig:greenhouse_a_cvcsfr_baseline}
  \end{minipage}

  \caption{Result of the baseline methods}
  \label{fig:baseline_result}

\end{figure*}

Fig. \ref{fig:baseline_result} shows qualitative results
of the UDA baseline methods on Greenhouse A dataset.
As shown quantitatively in \ref{sec:result-result-comparison},
the proposed method (CV+CS+FR) outperformed the baseline methods,
especially Seg-Uncertainty \cite{Zheng2021} and ProDA \cite{Zhang2021}.
In the segmentation results of Seg-Uncertainty and ProDA,
a large part of the bottom of the images are
wrongly classified as ground regions.
This is similar to the structural feature of the images in CS.
This tendency was possibly enhanced by the adversarial learning
on the output layer employed in the first stage of those methods.

Compared to Seg-Uncertainty and ProDA,
CRST \cite{Zou2019} produced better segmentation results.
CRST chooses pseudo-labels simply based on
the confidence of the predictions, and
resulting pseudo-labels are sparse.
In other words, CRST conservatively generates pseudo-labels
with only confidently predicted labels.
As a result, the performance was better than the other two baselines.
Our proposed method also generates pseudo-labels in a conservative way
via multiple model's agreement.
It may imply that it is a better approach to generate pseudo-labels
in a conservative way when the source and the target have
a large structural difference.
The proposed method provides a natural way of doing so
without relying on a specific dataset,
which allows for avoiding biased training.

\subsection{Effect of GAN-based image style transfer} 
\label{sec:GAN}


We trained the CycleGANs using CS and FR as source datasets and
Greenhouse A as a target dataset.
It is the first step of MADAN \cite{Zhao2019a},
a multi-source UDA method for semantic segmentation.
For each source dataset, we trained a CycleGAN for 200 epochs.

\begin{figure}[tb]
  \centering
  \begin{minipage}{0.24\hsize}
    \centering
    \includegraphics[width=\hsize]{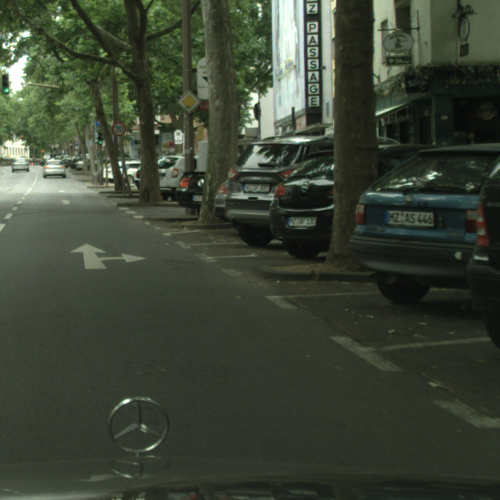}
    \subcaption{Real CS}
    \label{fig:cyclegan_examples_real_city}
  \end{minipage}
  \begin{minipage}{0.24\hsize}
    \centering
    \includegraphics[width=\hsize]{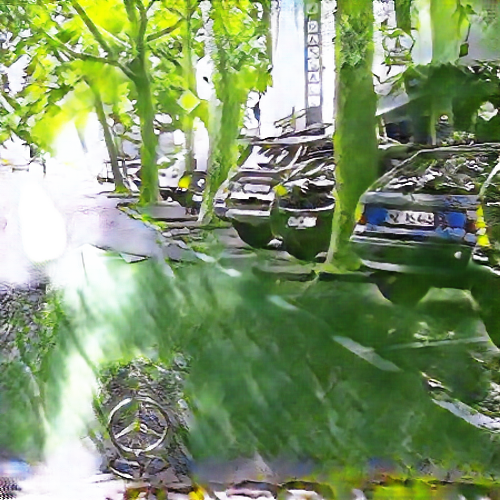}
    \subcaption{CS $\rightarrow$ Greenhouse}
    \label{fig:cyclegan_examples_fake_greenhouse}
  \end{minipage}
  \begin{minipage}{0.24\hsize}
    \centering
    \includegraphics[width=\hsize]{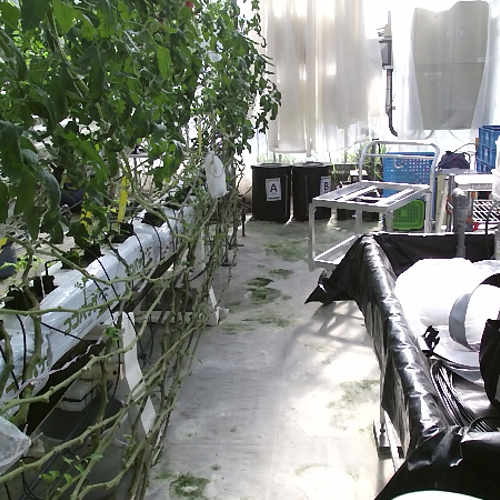}
    \subcaption{Real Greenhouse}
    \label{fig:cyclegan_examples_real_greenhouse}
  \end{minipage}
  \begin{minipage}{0.24\hsize}
    \centering
    \includegraphics[width=\hsize]{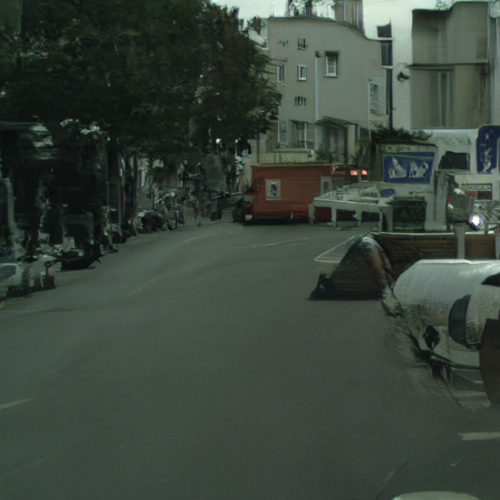}
    \subcaption{Greenhouse $\rightarrow$ CS}
    \label{fig:cyclegan_examples_fake_city}
  \end{minipage} \\

  \begin{minipage}{0.24\hsize}
    \centering
    \includegraphics[width=\hsize]{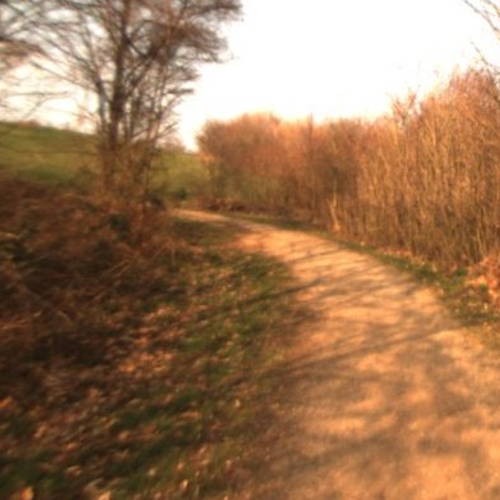}
    \subcaption{Real FR}
    \label{fig:cyclegan_examples_real_forest}
  \end{minipage}
  \begin{minipage}{0.24\hsize}
    \centering
    \includegraphics[width=\hsize]{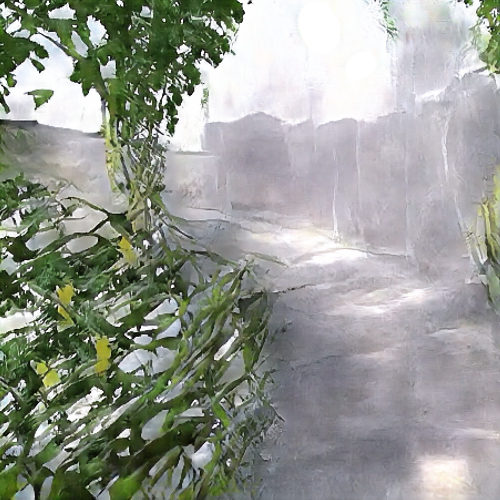}
    \subcaption{FR $\rightarrow$ Greenhouse}
    \label{fig:cyclegan_examples_fake_greenhouse_fr}
  \end{minipage}
  \begin{minipage}{0.24\hsize}
    \centering
    \includegraphics[width=\hsize]{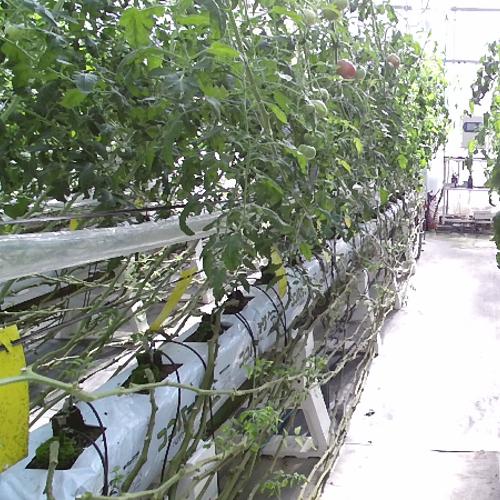}
    \subcaption{Real Greenhouse}
    \label{fig:cyclegan_examples_real_greenhouse_fr}
  \end{minipage}
  \begin{minipage}{0.24\hsize}
    \centering
    \includegraphics[width=\hsize]{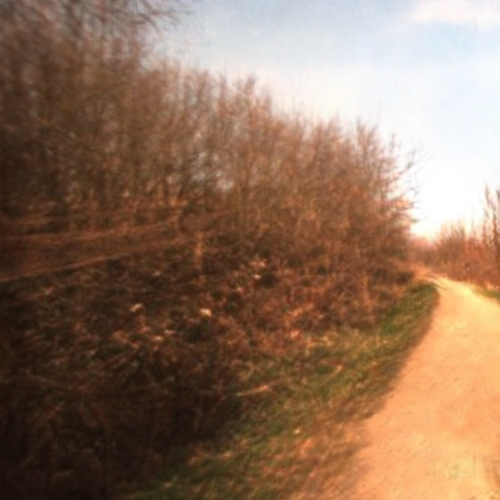}
    \subcaption{Greenhouse $\rightarrow$ FR}
    \label{fig:cyclegan_examples_fake_forest}
  \end{minipage}
  \caption{Results of image style transfer by CycleGAN in MADAN.}
  \label{fig:cyclegan_examples}
\end{figure}

Fig. \ref{fig:cyclegan_examples} shows the results of
image style transfer between CS/FR and Greenhouse A datasets.
The style transfer resulted in inconsistency of image contents,
e.g., the ground region in Fig. \ref{fig:cyclegan_examples_real_city}
is transferred to green plant-like objects in Fig. \ref{fig:cyclegan_examples_fake_greenhouse},
and the bottom part of the plant row and a large part of
an artificial object on the right in Fig. \ref{fig:cyclegan_examples_real_greenhouse}
are transferred to ground in Fig. \ref{fig:cyclegan_examples_fake_city}.
Similarly, plants on the right in Fig. \ref{fig:cyclegan_examples_real_forest}
are transferred to gray objects like a wall in Fig. \ref{fig:cyclegan_examples_fake_greenhouse_fr},
and some plant parts in Fig. \ref{fig:cyclegan_examples_real_greenhouse_fr}
are mapped to sky in \ref{fig:cyclegan_examples_fake_forest}.
This may be due to the large domain shift that stems from
the structural differences of the environments.

Although MADAN further employs training procedures to
aggregate the data from multiple domains closer to each other
with constraints on the semantic consistency \cite{Zhao2019a},
the inconsistency of the style transfer in the first step
shown above will affect the adaptation performance.

\end{document}